\documentclass[lettersize,journal]{IEEEtran}
\usepackage{amsmath,amsfonts}
\usepackage{array}
\usepackage[caption=false,font=normalsize,labelfont=sf,textfont=sf]{subfig}
\usepackage{textcomp}
\usepackage{stfloats}
\usepackage{url}
\usepackage{verbatim}
\usepackage{graphicx}
\usepackage{cite}
\usepackage{graphicx} 
\usepackage[utf8]{inputenc} 
\usepackage[T1]{fontenc}    
\usepackage{url}            
\usepackage{booktabs}       
\usepackage{amsfonts}       
\usepackage{nicefrac}       
\usepackage{microtype}      
\usepackage[dvipsnames]{xcolor}         
\usepackage[top=2cm,bottom=2cm,left=3cm,right=2.5cm]{geometry}
\usepackage{subfig}
\usepackage{floatrow}
\usepackage{amsmath,mathtools,amsthm}
\usepackage{color}
\usepackage{enumitem}
\usepackage{rotating}
\usepackage{multirow}
\usepackage[usestackEOL]{stackengine}
\usepackage{amssymb}
\usepackage[ruled]{algorithm2e}
\usepackage[noend]{algpseudocode}
\usepackage{booktabs}
\usepackage{array}

\usepackage{amsthm}
\usepackage[normalem]{ulem}
\usepackage{soul}
\mathtoolsset{showonlyrefs,showmanualtags}

\usepackage{mathtools}
\DeclarePairedDelimiter{\ceil}{\lceil}{\rceil}

\usepackage{bbm}

\newtheorem{theorem}{Theorem}
\newtheorem*{remark}{Remark}

\usepackage{placeins}
\newtheorem{proposition}{Proposition}
\newtheorem{lemma}{Lemma}
\usepackage{xargs}                      

\usepackage{xr-hyper}
\usepackage{hyperref}       

\makeatletter
\newcommand*{\addFileDependency}[1]{
  \typeout{(#1)}
  \@addtofilelist{#1}
  \IfFileExists{#1}{}{\typeout{No file #1.}}
}
\makeatother

\usepackage{amsthm}
\usepackage[normalem]{ulem}
\usepackage{soul}
\usepackage{bm}
\usepackage{authblk}

\title{Semi-Supervised Risk Control via Prediction-Powered Inference}

\author[1]{Bat-Sheva Einbinder}
\author[2]{Liran Ringel}
\author[1,2]{Yaniv Romano}
\affil[1]{Department of Electrical and Computer Engineering, Technion IIT, Haifa, Israel}
\affil[2]{Department of Computer Science, Technion IIT, Haifa, Israel}

\begin{document}

\maketitle

\begin{abstract}
    
The risk-controlling prediction sets (RCPS) framework is a general tool for transforming the output of any machine learning model to design a predictive rule with rigorous error rate control.
    The key idea behind this framework is to use labeled hold-out calibration data to tune a hyper-parameter that affects the error rate of the resulting prediction rule. However, the limitation of such a calibration scheme is that with limited hold-out data, the tuned hyper-parameter becomes noisy and leads to a prediction rule with an error rate that is often unnecessarily conservative. To overcome this sample-size barrier, we introduce a semi-supervised calibration procedure that leverages unlabeled data to rigorously tune the hyper-parameter without compromising statistical validity. Our procedure builds upon the prediction-powered inference framework, carefully tailoring it to risk-controlling tasks. We demonstrate the benefits and validity of our proposal through two real-data experiments: few-shot image classification and early time series classification.
\end{abstract}


\section{Introduction}
\IEEEPARstart{N}{owadays}, we are witnessing the widespread use of machine learning (ML) systems across various tasks. The remarkable potential of ML has led to its deployment in high-stakes applications such as autonomous driving and healthcare~\cite{rudin2019stop,tkatchenko2020machine,schneider2019mind,grigorescu2020survey}. However, in these critical tasks, it is crucial to enhance the reliability of predictive inference, as erroneous predictions can be harmful to users and society. One way to enhance the reliability of ML systems is to support their predictions with rigorous control over the expectation of a user-chosen loss function, known as the risk. 

Consider, for example, multi-class classification problems. Instead of reporting the most likely prediction of the classifier for a given test instance, it is often desired to return an uncertainty set of plausible labels, ensuring this set contains the unknown test label with high probability. In this scenario, we define a binary 0-1 loss function that checks whether the test label is included in the set. Consequently, the risk function---i.e., the expected value of the loss over test samples, is the miscoverage rate. Conformal prediction~\cite{papadopoulos2002inductive,lei2014distribution,angelopoulos2023conformal,balasubramanian2014conformal,vovk2005algorithmic} and its probably approximately correct (PAC) variant~\cite{parkpac} are capable of constructing statistically valid and calibrated uncertainty sets, controlling the miscoverate rate below a user-specified level, e.g., of 10\%. These frameworks have been further generalized to control other notions of error, such as the False Negative Rate (FNR) in multi-label tasks~\cite{bates2021distributionfree, angelopoulosconformal}, where the method we build upon in our work is the \textit{risk-controlling prediction sets} (RCPS)~\cite{bates2021distributionfree}. 

The calibration methods mentioned above share a common idea: given a prediction rule function, use a hold-out calibration data to tune a hyper-parameter that controls the risk for future samples. Under the assumption that the calibration and test data are i.i.d., these methods are guaranteed to control the risk in finite samples, for any data distribution and predictive model. Notably, the RCPS framework \cite{bates2021distributionfree} among other calibration methods \cite{papadopoulos2002inductive,lei2014distribution,angelopoulos2023conformal,balasubramanian2014conformal,vovk2005algorithmic,parkpac} assume that the risk is monotonous with respect to the hyper-parameter, meaning the risk decreases as the hyper-parameter increases. This assumption has been relaxed by the \emph{learn then test} approach~\cite{angelopoulos2025learn}, offering a rigorous framework that also uses hold-out calibration data but is capable of controlling more general risk functions that are non-monotonic with respect to the hyper-parameter that should be tuned.

While reliance on hold-out labeled data is crucial to attain distribution-free guarantees, it comes with a cost: when calibration methods are implemented with a small calibration data set they result in unnecessarily conservative risk control with high variability~\cite{vovk2005algorithmic,Vovk13,lei2014distribution,foygel2021limits,gibbs2025conformal,ding2024class,bates2023testing,angelopoulos2023conformal, lee2024batch}. This variability leads to divergent outcomes across multiple runs of the procedure due to the random choice of the limited calibration data. In practice, higher stability is desirable, as randomized methods tend to be less reliable and harder to interpret~\cite{bashari2024derandomized,yu2020veridical,murdoch2019definitions}. Limited data can occur when seeking risk control for a specific sub-population, such as individuals within specific regions of the covariate space defined by age, race, etc.~\cite{romano2020malice,tibshirani2019conformal,jung23multivalid,gibbs2025conformal,balasubramanian2014conformal,deng2023happymap,vovk2005algorithmic}. In this case, the underlying idea is to apply the calibration procedure to representative samples belonging to the same subgroup of the test point~\cite{jung23multivalid,gibbs2025conformal}. As such, the natural limitation here is that as the calibration becomes more localized, the sample size decreases. 
Few-shot learning is another example of having limited data for calibration, as only a few labeled samples are available by design~\cite{fisch2021few,bottou2007tradeoffs,wang2020generalizing}. 

The above discussion highlights that in many realistic settings, the problem of having access to limited labeled data is inevitable. This challenge leads us to the main question we ask in this work: \emph{how can we utilize unlabeled data to break sample size barriers?} Notably, the idea of using unlabeled data to improve the model's accuracy via semi-supervised learning is widely used in the machine learning literature. Utilizing unlabeled data is appealing since it mitigates the need for labeling efforts, making it more cost-effective and accessible; see, e.g.,~\cite{yang2022survey,hendrycks2019using,rizve2021defense}.
Nonetheless, this concept is under-explored for improving the sample efficiency of calibration procedures \cite{ding2024class,fisch2021few,seedat2023improving}, perhaps due to the challenge of the need to provide rigorous statistical guarantees for out-of-sample data.

In this work, we build upon \emph{prediction-powered inference} (PPI)~\cite{angelopoulos2023prediction}, a semi-supervised framework that constructs valid confidence intervals for statistical parameters of a distribution, such as their mean or quantiles. We show how this framework can be utilized to enhance the sample efficiency of risk-controlling techniques, all without compromising the correctness of the statistical assurance.
The key idea is to use the imputed labels of the unlabeled data to tune the hyper-parameter that affects the risk function and then correct the tuning process with the few labeled data, all while rigorously accounting for the inaccuracies in the imputed labels. As a result, our method reduces the variability of the calibration process when the imputed labels are accurate while maintaining validity even if the imputations are incorrect. This stands in contrast to a naive approach that directly augments the labeled and unlabeled data, which can lead to invalid calibration when the imputed labels are inaccurate.

To summarize, our contributions are the following:
\begin{itemize}
    \item We first prove that vanilla RCPS procedure controls non-monotonic risk functions and not merely monotonic risks as shown in the original paper~\cite{bates2021distributionfree}. While our proof builds on the foundations of~\cite{angelopoulos2025learn}, RCPS is arguably more intuitive and interpretable than \cite{angelopoulos2025learn}, as the latter requires a more comprehensive understanding and background on multiple hypotheses testing.
    This result is of independent interest and unlocks the application of our semi-supervised calibration procedures to any risk function.
    \item Then, we proceed to our main contribution: the proposal of a semi-supervised calibration procedure that leverages unlabeled data to improve sample efficiency while ensuring statistical validity. Our method can be applied to any risk function, including non-monotonic risks. Additionally, we introduce a specialized calibration approach for binary losses which are very common in practice. This specialized procedure carefully utilizes the binary property of the loss function, resulting in further enhancements.
    \item To demonstrate the validity and advantages of our methods we conduct two real data experiments. The first application deals with few-shot calibration in image classification using ImageNet dataset. Here, we construct prediction sets that control the miscoverage risk. We show that our method tends to reduce the conservatism and variability of the coverage, especially when the imputations are reasonably accurate. The second application involves early exit classification in a natural language processing (NLP) task, aiming to reliably terminate the inference process of a large language model (LLM) while controlling the accuracy gap between full and partial text scanning. We show that using unlabeled data can promote early exits without compromising validity.
\end{itemize}

\section{Background and Problem Setup}
Denote by $D^{\text{cal}}_L=\{(X_i,Y_i)\}_{i=1}^n$ a labeled calibration set such that $(X_i,Y_i)$ are sampled i.i.d. from $P_{XY}$ for all $1\leq i \leq n$. The feature $X$ takes values in $\mathcal{X}$ and the response $Y$ takes values in $\mathcal{Y}$. Given $X$, let $\mathcal{T}_{q}(X)$ denote a post-processed prediction function for $Y$ based on a pre-trained model $\hat{f}$, where the associated risk is controlled by a tunable hyperparameter $q$. For example, in a classification task, $\mathcal{T}_q(X)$ is a prediction set that can be constructed by including all labels whose estimated class probabilities pass the threshold $1-q$. Finally, let $L(Y,\mathcal{T}_q(X))$ be a loss function that measures an error notion between the true response $Y$ and $\mathcal{T}_q(X)$; its expectation over future test samples $(X,Y)$, sampled i.i.d. from $P$ is the risk function, i.e., $R(q)=\mathbb{E}[L(Y,\mathcal{T}_q(X))]$. 
In the classification example, the loss could be the binary miscoverage loss $L(Y,\mathcal{T}_q(X))=\mathbbm{1}\{Y\notin \mathcal{T}_q(X)\}$, where $\mathbbm{1}\{A\}$ is the indicator function that equals one if $A$ occurs, and zero otherwise.

Focusing on classification settings, the work reported in~\cite{parkpac} offers a framework to construct PAC prediction sets, aiming to find $\hat{q}$ that satisfies the following statement under the i.i.d. assumption:
\begin{equation}
\label{eq:PAC-sets}
    \mathbb{P}_{D^{\text{cal}}_L}(\mathbb{E}[\mathbbm{1}\{Y\notin \mathcal{T}_{\hat{q}}(X)\}] \leq \alpha) \geq 1-\delta,
\end{equation}
where $(\alpha,\delta)$ are user-specified error levels. The above states that with probability $(1-\delta)$ the miscoverage rate of the set-valued prediction rule $\mathcal{T}_{\hat{q}}$ will fall below $\alpha$. Here, the probability is taken over the calibration data, which is used to tune the hyper-parameter $\hat{q}$.

The RCPS method \cite{bates2021distributionfree} generalizes \cite{parkpac} to situations where one is interested in controlling any monotonic risk function at levels $(\alpha,\delta)$:
\begin{equation}
\label{eq:rcps}
    \mathbb{P}_{D^{\text{cal}}_L}(R(\hat{q}) \leq \alpha) \geq 1-\delta.
\end{equation}
Monotonicity means the risk decreases as the parameter $q$ increases. This occurs, for example, when predicting sets that are nested, i.e., $q_1\geq q_2 \Rightarrow \mathcal{T}_{q_1}(X)\supseteq \mathcal{T}_{q_2}(X)$ and $\mathcal{T}_{q_1}(X)\supseteq \mathcal{T}_{q_2}(X) \Rightarrow L(Y,\mathcal{T}_{q_1}(X))\leq L(Y,\mathcal{T}_{q_2}(X))$. 

In general, however, the monotonicity property does not always hold. For example, consider a situation where the response is high-dimensional, as happens in multi-label classification, and we wish to construct a set-valued prediction rule $\mathcal{T}_q(X)$ that controls the False Discovery Rate (FDR), which is a non-monotonous risk with respect to $q$. To address this issue, the RCPS framework was further generalized in \cite{angelopoulos2025learn}, allowing the control of non-monotonic and even multiple risks. The idea is to reformulate the risk-control task as a multiple hypothesis testing problem, testing which of the different values of the hyper-parameter $q$ rigorously controls the risk. Our first contribution reveals that the RCPS framework also controls non-monotonic risks, where our proof builds on the same concepts from \cite{angelopoulos2025learn}.

In our work, we aim to further expand RCPS to leverage unlabeled data, enabling more efficient tuning of $\hat{q}$.
Denote further by $D^{\text{cal}}_U=\{X_j\}_{j=n+1}^{n+N}$ a large independent unlabeled calibration set such that  $\{X_j\}_{j=n+1}^{n+N}$ and $\{X_i\}_{i=1}^n$ are i.i.d. Also, denote the imputed labels---obtained from a fixed predictive model---of a sample $X_j$ by $\tilde{Y}_j$.
Using both labeled and unlabeled calibration sets we aim to tune $\hat{q}$ to achieve:
\begin{equation}
    \label{eq:pac-coverage-new}
    \mathbb{P}_{D^{\text{cal}}_U,D^{\text{cal}}_L}(R(\hat{q})\leq \alpha)\geq 1-\delta,    
\end{equation} 
where, in contrast to~\eqref{eq:rcps}, the probability is taken over the randomness in the labeled and unlabeled calibration 
data.



\section{Related Work}
\label{sec:related_work}

Recently, a growing body of work has focused on semi-supervised estimation of statistical objectives using many unlabeled data points and few labeled ones~\cite{angelopoulos2023prediction,chakrabortty2022semi,hou2023surrogate,chen2005measurement,wu2001model,carroll2006measurement,Chakrabortty_2018,zhang2022high,zhu2024doubly,angelopoulos2023ppi++,zrnic2024cross,masserano2023simulator,lu2024quantifying}. In our paper, we build on the PPI framework~\cite{angelopoulos2023prediction}: a semi-supervised estimation method for constructing valid confidence intervals for a statistical quantity of interest, such as the mean or a quantile of a distribution. The PPI method is supported by both asymptotic and non-asymptotic guarantees for the coverage of the population object of interest. Its appeal is that the variance of the estimator can be reduced when an accurate predictive model is available. However, constructing these confidence intervals with finite sample guarantees often leads to overly conservative results. This emphasizes the need to tailor this method to the important problem of constructing risk-controlling prediction sets, and leveraging properties of the loss for further enhancements. 

In the context of conformal prediction, there have been attempts to overcome sample size barriers, however, all these attempts do not include using unlabeled data. In~\cite{fisch2021few}, the authors extend conformal prediction to include few-shot predictions with auxiliary tasks and create a meta-learning framework for constructing confident set-valued classifiers for new target tasks. Put simply, their approach leverages different but related tasks to enhance sample efficiency. This stands in contrast with our method which leverages unlabeled data. As such, one can view our method as an orthogonal direction to improve the sample efficiency of calibration methods.
In~\cite{ding2024class}, the authors focus on class-conditional coverage guarantees of conformal methods and suggest utilizing similarities between different classes to enhance the sample size of the calibration data for each class. Here, sample size issues become apparent when the number of classes is large, as the calibration set for each class becomes small. To mitigate this issue, the authors introduced a clustering approach, which combines samples from different classes that have similar prediction errors. In turn, this method attains a cluster-conditional rather than class-conditional coverage.
Other approaches to improve statistical efficiency include cross-validation~\cite{cohen2024cross}, meta-learning~\cite{park2022pac,park2023few}, and more efficient use of data for model- and score-selection~\cite{sharma2023pac}.

Lastly, independent work by~\cite{xu2024active}, posted concurrently with our proposal, extends RCPS to sequential settings with an active labeling mechanism, ensuring valid risk control guarantees across all time steps. Similar to this paper, the authors of~\cite{xu2024active} also leverage PPI for variance reduction; however, their setup differs significantly from ours. First, the method in \cite{xu2024active} aims to reduce variance per input to improve the effectiveness of active labeling, but this approach would have no impact in the offline setting we consider in this paper, as it does not involve active labeling. Second, the validity of~\cite{xu2024active} only holds for monotonous risks, an assumption we do not make. In fact, it would be illuminating to further explore whether our method can extend their approach to handle non-monotonic risks.

\section{Warm Up: Non-Monotonic Risk Control via RCPS}
\label{sec:warm_up}


Since our approach builds upon the RCPS framework, we describe it here in detail. Recall the risk $R(q)=\mathbb{E}[L(Y,\mathcal{T}_q(X))]$ where $q$ is a parameter that should be rigorously tuned to satisfy~\eqref{eq:rcps}.
This tuning is done by using the hold-out data and a proper concentration inequality to find an upper confidence bound (UCB) $\hat{R}^+(q)$ for $R(q)$ with error level $\delta$:
\begin{equation}
\label{eq:bound}
\mathbb{P}_{D^{\text{cal}}_L}(R(q)\leq\hat{R}^+(q))\geq 1-\delta \hspace{2mm} \forall q.
\end{equation}
There are several ways to construct such a UCB. For example, we can apply Hoeffding's inequality~\cite{hoeffding1994probability} and get
$$\hat{R}^+_{\text{Hoef}}(q)=\hat{R}(q)+\sqrt{\frac{1}{2n}\log{\frac{1}{\delta}}},$$
where $\hat{R}(q)=\frac{1}{n}\sum_{i=1}^{n}L(Y_i,\mathcal{T}_q(X_i))$ is the empirical risk. Other methods for constructing a UCB include Waudby-Smith–Ramdas (WSR) bound~\cite{waudby2024estimating}, Hoeffding–Bentkus bound~\cite{bentkus2004hoeffding, bates2021distributionfree}, Clopper–Pearson bound~\cite{clopper1934use}, and more.

Denote by $q_{\text{max}}$ the smallest real value for which the set size is infinite, and by $q_{\text{min}}$ the largest real number for which the set is empty. Armed with a method to construct a UCB that satisfies~\eqref{eq:bound}, the RCPS procedure iterates over all possible values of $q$ from $q_{\text{max}}$ to $q_{\text{min}}$ in decreasing order and returns the first value $\hat{q}$ for which the bound is not less than $\alpha$. 
Formally,
\begin{equation}
\label{eq:find_q}
    \hat{q} \triangleq \inf\{q\in \mathcal{Q}_{\text{grid}}:\hat{R}^+(q')<\alpha, \hspace{2mm} \forall q'>q\},
\end{equation}
where $\mathcal{Q}_{\text{grid}}=\{q_{\text{max}},q_{\text{max}}-\Delta, q_{\text{max}}-2\Delta,\ldots q_{\text{min}}\}$ and $\Delta$ is some small constant. When the risk is monotonous with respect to $q$, as assumed in~\cite{bates2021distributionfree}, this grid indeed leads to a sensible $\hat{q}$, since the risk increases as $q$ decreases. Importantly, it is proven in~\cite{bates2021distributionfree} that this calibration procedure satisfies~\eqref{eq:rcps}.

Herein, we present a new result on the validity of RCPS, indicating that this framework in fact controls any general non-monotonic risk function and not merely monotonic ones, as argued in \cite{bates2021distributionfree}. The procedure is similar to RCPS, however the pre-specified grid $\mathcal{Q}_{\text{grid}}$ might differ, especially when we would not expect the risk to be small for larger $q$ values. Here, the choice of $\mathcal{Q}_{\text{grid}}$ is crucial as, by design, the procedure stops the first time $\hat{R}^+(q)$ is not below $\alpha$. As such, when the risk is not monotonous, the grid $\mathcal{Q}_{\text{grid}}$ should be carefully designed to increase statistical efficiency~\cite{ringel2024early,angelopoulos2025learn,lauferefficiently}. 
For example, our early time series classification experiment in Section~\ref{sec:exp_nlp} involves tuning a complex, high-dimensional hyperparameter vector. To do so, we adopt the approach proposed in~\cite{ringel2024early}, which is related to the methods introduced in~\cite{lauferefficiently,lauferrisk}. 
With this design choice in mind, let $\mathcal{Q}_\text{grid}=\{q_1,q_2,\ldots, q_M\}$ be a pre-specified grid for the parameter $q$; in the monotonous case, $q_1=q_{\text{max}}, q_2=q_{\text{max}}-\Delta,\ldots$ Then, the procedure proceeds by computing a UCB with error level $\delta$ at these specified points $(\hat{R}^+(q_1),\hat{R}^+(q_2),\ldots, \hat{R}^+(q_M))$. We iterate over this grid in order, from $q_1$ to $q_M$, and return the first value for which the UCB is not less than $\alpha$, as in~\eqref{eq:find_q}.

\begin{theorem}
\label{thm:fst}
    Assuming the calibration and test set are i.i.d. and given an upper confidence bound that satisfies~\eqref{eq:bound}, the RCPS procedure satisfies~\eqref{eq:rcps} for any general risk function (including non-monotonous risks).
\end{theorem}

This result indicates that the regular RCPS algorithm can be used to control any risk function in finite samples, for any data distribution, and regardless of the choice of the predictive model $\hat{f}$ used to construct $
\mathcal{T}_{q}(X)$. This conclusion is not only of independent interest, but also important for the derivation of the main contribution of this work: a semi-supervised calibration procedure with rigorous risk control. 

We pause to explain the high-level idea of the proof of Theorem~\ref{thm:fst}. It is based on the \emph{learn then test} framework presented in~\cite{angelopoulos2025learn} which expands RCPS to control also non-monotonous risk functions.
It reformulates risk control as a multiple-hypothesis testing task, testing for each $q \in \mathcal{Q}_{\text{grid}}$ whether the risk is not controlled. As such, the null hypothesis for a given $q$ is $H_0: \ R(q)>\alpha$. 
The idea is to iterate over $\mathcal{Q}_{\text{grid}}$, calculate a p-value for each hypothesis, and stop the first time the null is not rejected. 
We show that RCPS and \emph{learn then test} are equivalent when implemented with a search grid $\mathcal{Q}_\text{grid}= \{q_1,q_2,...,q_M\}$ of the same ordering. This is due to the duality between p-values and confidence bounds. 
As a result, the validity of RCPS for non-monotonous risks follows naturally. 
The detailed algorithm and proof are provided in Appendix~\ref{app:fst}.

\section{Semi-Supervised RCPS: A General Loss}
\label{sec:ss_RCPS_general}

We now turn to introduce our approach to leverage unsupervised data to enhance the sample efficiency of RCPS, while preserving its rigorous distribution-free validity guarantees that holds in finite samples.
To formalize the procedure, denote by $\{L_i(q)\}_{i=1}^n$ and $\{\tilde{L}_j(q)\}_{j=1}^{n+N}$ the labeled and pseudo-labeled losses. The labeled loss  $L_i(q)=L(Y_i,\mathcal{T}_q(X_i))$, $i=1,\ldots,n$, is evaluated on the small labeled calibration set, the loss $\tilde{L}_j(q)=L(\tilde{Y}_j,\mathcal{T}_q(X_j))$, $j=n+1,\ldots,N+n$ is evaluated on the unlabeled data, and $ \tilde{L}_i(q)=L(\tilde{Y}_i,\mathcal{T}_q(X_i))$, $i=1,\ldots,n$ is an additional loss derived from the labeled data but using imputed labels.

In~\cite{angelopoulos2023prediction}, the authors introduce the prediction-powered risk, defined as follows:
\begin{equation}
   \label{eq:PP-risk}
    \hat{R}_\text{PP}(q)=\underbrace{\frac{1}{N}\sum_{j=n+1}^{n+N} \tilde{L}_j(q)}_{\hat{R}_U(q)}+\underbrace{\underbrace{\frac{1}{n}\sum_{i=1}^n L_i(q)}_{\hat{R}_L(q)}-\frac{1}{n}\sum_{i=1}^n \tilde{L}_i(q)}_{\hat{R}_{\text{rect}}(q)}.
\end{equation}
The first term, $\hat{R}_U(q)$ is the empirical risk evaluated only with the unlabeled imputed data, and $\hat{R}_{\text{rect}}(q)$ is the \textit{rectifying} risk which estimates the bias introduced by the imputation. It is shown in~\cite{angelopoulos2023prediction} that~\eqref{eq:PP-risk} is an unbiased estimator of $R(q)$, regardless of the accuracy of the imputed labels $\{\tilde{Y}\}_{i=1}^n$ and $\{\tilde{Y}\}_{j=n+1}^{n+N}$. To see this, we can take the expectation over the labeled and unlabeled data, and achieve:
\begin{align}
    R_{\text{PP}}(q)=\mathbb{E}[\hat{R}_\text{PP}(q)]&=\underbrace{\mathbb{E}[\hat{R}_U(q)]}_{R_U(q)}+\underbrace{\mathbb{E}[\hat{R}_{\text{rect}}(q)]}_{R_{\text{rect}}(q)}\\
    &=\mathbb{E}[L(q)]=R(q).
    \label{eq:unbiased}
\end{align}
This is because the imputed losses are identically distributed, thus
\begin{align}
    &\mathbb{E}\left[\frac{1}{N}\sum_{j=n+1}^{n+N} \tilde{L}_j(q) - \frac{1}{n}\sum_{i=1}^n \tilde{L}_i(q)\right] \\
    &=\mathbb{E}\left[\tilde{L}_j(q)\right]-\mathbb{E}\left[\tilde{L}_i(q)\right]=0, 
\end{align}
leading to $$\mathbb{E}\left[\hat{R}_\text{PP}(q)\right] = \mathbb{E}\left[\frac{1}{n}\sum_{i=1}^n L_i(q)\right]=\mathbb{E}\left[L(q)\right]=R(q).$$

Next, we explain when and why this risk reduces the estimator's variance compared to using only labeled data for constructing confidence intervals or upper confidence bounds. The variance of the empirical risk $\hat{R}_L(q)$ that uses only labeled data is $\hat{\sigma}^2_L={{\sigma}_{L_i}^2}/{n}$, where ${\sigma}_{L_i}^2$ is the variance of $L_i(q)$, which we cannot control.
On the other hand, the variance of the empirical prediction-powered risk $\hat{R}_{\text{PP}}(q)$ in~\eqref{eq:PP-risk} is $\hat{\sigma}^2_{\text{PP}}=\hat{\sigma}^2_U+\hat{\sigma}^2_{\text{rect}}={{\sigma}^2_{\tilde{L}_j}}/{N}+{{\sigma}_{\tilde{L}_i-L_i}^2}/{n}$. Here, ${\sigma}^2_{\tilde{L}_j}$ is the variance of $\tilde{L}_j(q)$ and ${\sigma}_{\tilde{L}_i-L_i}^2$ is the variance of $\tilde{L}_i(q)-L_i(q)$. Notably, the variance of $\hat{R}_{\text{PP}}(q)$ is the sum of these two variances, since $\{\tilde{L}_i(q)-L_i(q)\}_{i=1}^n$ and $\{\tilde{L}_j(q)\}_{j=n+1}^{n+N}$ are evaluated on independent samples.
With a large number $N$ of unlabeled samples, ${\sigma^2_{\tilde{L}_j}}/{N}$ is close to zero. Moreover, $\sigma_{\tilde{L}_i-L_i}^2$ becomes smaller as the imputation accuracy increases. 
In turn, when the imputed labels are reasonably accurate, $\hat{\sigma}_{\text{PP}}^2$ is anticipated to be smaller than $\hat{\sigma}^2_L$. Another way to view the benefit of PPI is this: when the imputations are accurate, $\hat{R}_{\text{rect}}(q)$ in \eqref{eq:PP-risk} is approximately zero, while $\hat{R}_U(q)$ is close to the true risk as the sample size $N$ is large. This, in turn, can be beneficial to obtain a tighter UCB, as we show next. 
Nonetheless, even with inaccurate imputations, the UCB will remain valid as $\hat{R}_{\text{PP}}(q)$ is unbiased, though we would not gain improvement in variance reduction compared to using only labeled data, and might even lose statistical efficiency. At the end of this section and in Appendix~\ref{app:baselines_lambda}, we discuss a method that follows~\cite{angelopoulos2023ppi++} and mitigates this negative effect by introducing a hyperparameter to control the influence of unlabeled data, thereby enabling more controlled variance reduction in the PPI risk estimator.

To attain finite-sample risk control, we shall use concentration inequalities to compute a UCB using the prediction-powered risk. To do so, we first re-write the risk in a way that breaks the sum in \eqref{eq:PP-risk} into a sum of $n$ i.i.d. random variables, as follows: 
\begin{align}
    &\hat{R}_\text{PP}(q) = \\ 
    &= \frac{1}{n}\sum_{i=1}^n\left(\frac{n}{N}\sum_{j=(i-1)\frac{N}{n}+n+1}^{i \frac{N }{n}+n} \tilde{L}_j(q) +L_i(q)-\tilde{L}_i(q) \right) \\
    &=\frac{1}{n}\sum_{i=1}^nW_i. \label{eq:ppi_decom}
\end{align}
Above, we split $\hat{R}_U(q)$ in \eqref{eq:PP-risk} into blocks of length ${N}/{n}$. If ${N}/{n}$ is not an integer, we can use the floor value, slightly reducing the effective number of unlabeled points. 
Therefore, given the i.i.d. random variables $\{ W_i \}_{i=1}^n$, we can invoke any concentration inequality to obtain a UCB for the risk ${R}_{\text{PP}}(q)={R}(q)$ that satisfies:
\begin{equation}
\label{eq:pp-bound}
\mathbb{P}_{D^{\text{cal}}_L, D^{\text{cal}}_U}(R(q)\leq\hat{R}_{\textup{PP}}^+(q))\geq 1-\delta \hspace{2mm} \forall q.
\end{equation}
In practice, we recommend using the state-of-the-art WSR UCB described in Algorithm~\ref{alg:wsr} of Appendix~\ref{app:wsr} to attain finite-sample validity or build upon the central limit theorem (CLT) to construct an asymptotically valid UCB. The latter can be less conservative, at the cost of relying on stronger assumptions. Notably, the random variables $\{W_i\}_{i=1}^n$ have different support from the original losses $L_i(q)$. 
For example, in the binary case, $L_i(q)\in[0,1],$ while $W_i\in[-1,2]$.
One should account for this discrepancy when constructing the UCB, especially because various concentration inequalities, such as WSR, assume the variables are bounded between $0$ and $1$. To tackle this, in Algorithm~\ref{alg:wsr} of Appendix~\ref{app:wsr} we present a simple extension of the original WSR for losses bounded between any two numbers $A$ and $B$, similar to~\cite{jazbecfast}. 
Note that linearly scaling the loss to the $[0,1]$ range is also possible. As demonstrated in Appendix~\ref{app:wsr}, such a linear scaling approach achieves comparable power to the generalized WSR UCB method, which extends beyond the standard $[0,1]$ loss.

For ease of reference, the detailed calibration procedure is depicted in Algorithm~\ref{alg:ss_calib_blocks}, and the following Proposition~\ref{thm:valid-ss-cal-blocks} states its validity. The proof follows directly from the validity of the UCB combined with Theorem~\ref{thm:fst}.

\begin{algorithm}[ht]
\caption{Semi-Supervised RCPS for a General loss}\label{alg:ss_calib_blocks}
\textbf{Input:} \\
Labeled calibration data $\mathcal{D}_{L}^{\text{cal}}=\{(X_i,Y_i)\}_{i=1}^{n}$; \\
Unlabeled calibration data $\mathcal{D}_{U}^{\text{cal}}=\{X_j\}_{j=n+1}^{n+N}$;\\
Imputed labels $\{\tilde{Y}_j\}_{j=1}^{n+N}$; \\
A set $\mathcal{Q}_\text{grid}$ containing a grid of values for $q$;\\
Loss function $L$;\\
Risk control levels $(\alpha,\delta)$. \\

\algnewcommand{\LeftComment}[1]{{\color{gray}\(\triangleright\) #1}}

\textbf{Process:}\\
        \For {$q$ in $\mathcal{Q}_\textup{grid}$}
        {For all $1\leq i\leq n$, compute $$W_i=\frac{n}{N}\sum_{j=(i-1)\frac{N}{n}+n+1}^{i\frac{N}{n}+n} \tilde{L}_j(q)+L_i(q)-\tilde{L}_i(q)$$ \\
        Derive a UCB $\hat{R}_{\text{PP}}^+(q)$ given $\hat{R}_\text{PP}(q)$ in \eqref{eq:ppi_decom} \\ \LeftComment{E.g., using WSR in Algorithm~\ref{alg:wsr} of Appendix~\ref{app:wsr}}\\
        \eIf{$\hat{R}_{\textup{PP}}^+(q)<\alpha$}
        {$\hat{q} \leftarrow q$}
        {break}}
        
\textbf{Output:} The calibrated parameter $\hat{q}$.
\end{algorithm}

\begin{proposition}
    \label{thm:valid-ss-cal-blocks}
 Under the assumption that the labeled calibration set $\{(X_i,Y_i)\}_{i=1}^n$ and the test data are i.i.d, and the unlabeled covariates $\{X_j\}_{j=n+1}^{n+N}$ are i.i.d with $\{X_i\}_{i=1}^{n}$, the calibrated parameter $\hat{q}$ constructed by Algorithm~\ref{alg:ss_calib_blocks} satisfies~\eqref{eq:pac-coverage-new}. 
\end{proposition}
\begin{remark}
The risk-controlling guarantee depends on the validity of the UCB. Specifically, if~\eqref{eq:pp-bound} holds in finite samples---e.g., when constructing $\hat{R}_{\textup{PP}}^+(q)$ via WSR---then~\eqref{eq:pac-coverage-new} would hold. In contrast, when using asymptotically valid UCB with CLT, the risk-controlling guarantee would only hold asymptotically.
\end{remark}

Note that the UCB constructed with the CLT is given by $\bar{W}+z_{1-\delta}\cdot \sqrt{\frac{\sigma^2_W}{n}}$ where $\bar{W}=\frac{1}{n}\sum_{i=1}^{n}W_i$, $z_{1-\delta}$ is the $1-\delta$ quantile of the standard normal distribution, and $\sigma^2_W$ is the variance of $W_i$. Since we assume the labeled and unlabeled points are i.i.d., the above CLT-based UCB is equivalent to a one-sided confidence interval constructed with Algorithm 1 of~\cite{angelopoulos2023prediction}. Thus, when resorting to asymptotic guarantees,
one can further enhance the statistical efficiency of PPI, as described in \cite{angelopoulos2023ppi++}. The idea is to introduce a hyper-parameter $\lambda\in\mathbb{R}$ that controls the reliance on the unlabeled data. This is done by multiplying by $\lambda$ the imputed losses $\frac{1}{N}\sum_{j=n+1}^{n+N} \tilde{L}_j(q) - \frac{1}{n}\sum_{i=1}^n \tilde{L}_i(q)$ in \eqref{eq:PP-risk}. This hyper-parameter $\lambda$ can be set in a data-driven fashion, where the higher the accuracy the larger the $\lambda$ will be. Nonetheless, $\lambda$ cannot be tuned when providing finite-sample guarantees without assuming access to additional hold-out data. For this reason, in our method, we set $\lambda=1$ by default. See Appendix~\ref{app:baselines_lambda} for further details and additional baselines involving this parameter.

Focusing on finite-sample guarantees, we should stress that the proposed decomposition of the prediction-powered risk in \eqref{eq:ppi_decom} offers a different and possibly more powerful approach to construct a valid UCB compared to the technique presented in~\cite{angelopoulos2023prediction}.
The approach in~\cite{angelopoulos2023prediction} requires splitting the error budget $\delta=\delta_1 + \delta_2$ between two UCBs: one corresponding to the unlabeled risk $\hat{R}_U(q)$ and the other corresponding to the rectifying risk $\hat{R}_{\text{rect}}(q)$.
As such, the final UCB of $R(q)$ is the sum of the two; we discuss this budget-splitting strategy in more detail in the next section.
Naturally, budget-splitting can reduce the statistical efficiency, especially for the rectifying risk $\hat{R}_{\text{rect}}(q)$ that involves a few samples, whose UCB must be obtained for $\delta_2 < \delta$. By contrast, our proposed approach does not require splitting the error budget. More broadly, our new formulation can also be used to construct finite-sample valid confidence intervals for a population parameter. This approach has the potential to produce smaller intervals than those presented in~\cite{angelopoulos2023prediction}, making it of independent interest.

There are special cases, however, in which budget-splitting can be beneficial---not due to the splitting itself, but because it preserves the properties of the loss. For example, when the loss $L(q)$ is binary the UCB can be derived precisely, as we discuss next. A closer look into $W_i$ reveals that this formulation breaks the binary property of the loss, preventing us from using the exact UCB. Importantly, binary losses are at the heart of miscoverage rate control, selective classification, and more~\cite{ringel2024early,park2020pac,parkpac,angelopoulos2025learn,lauferefficiently,angelopoulosconformal}.
For this reason, we turn to present a specialized procedure that preserves the binary property of the loss and uses an exact UCB.

\section{Specialized Semi-Supervised RCPS: a Binary Loss}
\label{sec:ss_RCPS_binary}

In situations where the loss is binary, the random variable $L(Y,\mathcal{T}_q(X))$ follows the Bernoulli distribution with success probability that is equal to the risk $R(q)=\mathbb{E}[L(Y,\mathcal{T}_q(X))]$. This observation implies that the empirical error count $\sum_{i=1}^n L_i(q)$ follows a binomial distribution $\text{Binom}(n,R(q))$ with $n$ independent trials and success probability of $R(q)$ for each trial. 
With this in place, we can use the exact finite-sample Clopper-Pearson UCB~\cite{clopper1934use} for the risk, formulated as: 
\begin{equation}
\label{eq:exact_bound}
    \hat{R}^+(q)=\sup\{R:\mathbb{P}(\text{Binom}(n,R)\leq\ceil{n\hat{R}_L(q)})\geq \delta\}.
\end{equation}
Notably, the Clopper-Pearson UCB is exact in the sense that it is based directly on the binomial distribution and does not use any approximation or concentration inequalities. This exactness is desired especially in our setting due to the limited number of labeled samples available.

To utilize this exact UCB in our semi-supervised setting, we first upper bound the prediction-powered risk~\eqref{eq:PP-risk} as follows:
\begin{align}
    R_{\text{PP}}(q) &=R_U(q)+R_{\text{rect}}(q) \\
    & \leq R_U(q)+R^{\text{clip}}_{\text{rect}}(q) = R_{\text{PP}}^\text{clip}(q), \label{eq:pp-clip}
\end{align}
where $R^{\text{clip}}_{\text{rect}}(q)=\mathbb{E}[(L(q)-\tilde{L}(q))_+].$
The operator $(z)_+$ returns $\max\{z,0\}$. While $(L(q)-\tilde{L}(q))$ can have values in $\{-1,0,1\}$, its clipped version $(L(q)-\tilde{L}(q))_+$ can only achieve values in $\{0,1\}$. In turn, the clipping enables us to use the exact Clopper–Pearson UCB for both $R^{\text{clip}}_{\text{rect}}(q)$ and $R_U(q)$, as the latter involves a binary loss by definition. 


In the calibration procedure, we split the error budget $\delta = \delta_1+\delta_2$, where $\delta_1,\delta_2 > 0$. We continue by calculating two separate UCBs: $\hat{R}^+_U(q)$ for $R_U(q)$ at level $\delta_1$, 
and $\hat{R}^{\text{clip}+}_{\text{rect}}(q)$ for  $R^{\text{clip}}_{\text{rect}}(q)$ at level $\delta_2$. These bounds can be constructed similarly to~\eqref{eq:exact_bound} with the empirical risks
\begin{equation}
\label{eq:emp_risks}
    \hat{R}_{U}(q) \hspace{2mm} \text{and} \hspace{2mm}
    \hat{R}^{\text{clip}}_{\text{rect}}(q)=\frac{1}{n}\sum_{i=1}^n(L_i(q)-\tilde{L}_i(q))_+,
\end{equation}
plugged into \eqref{eq:exact_bound} instead of $\hat{R}_L$. As a result, we have
\begin{align}
\label{2_bounds}
    &\mathbb{P}_{{D}^{\text{cal}}_U}(R_{U}(q)\leq \hat{R}^+_{U}(q))\geq 1-\delta_1, \hspace{2mm} \\
    &\mathbb{P}_{{D}^{\text{cal}}_L}(R^{\text{clip}}_{\text{rect}}(q)\leq \hat{R}^{\text{clip}+}_{\text{rect}}(q))\geq 1-\delta_2.
\end{align}
Due to the significant difference in size between the labeled and unlabeled datasets, we recommend setting $\delta_1 \ll \delta_2$. Finally, we construct $\hat{R}^+(q)$, the UCB of the risk $R(q)$, by $\hat{R}^+(q)=\hat{R}^+_{U}(q)+\hat{R}^{\text{clip}+}_{\text{rect}}(q)$. The following result states the validity of the bound $\hat{R}^+(q)$.
\begin{lemma} \label{thm:valid_bound}
    Under the assumptions of Proposition~\ref{thm:valid-ss-cal-blocks}, and given the UCBs $\hat{R}^+_{U}(q)$ and $\hat{R}^{\textup{clip}+}_{\textup{rect}}(q)$ that satisfy~\eqref{2_bounds}, then:
    $$\mathbb{P}_{{D}^{\textup{cal}}_L,{D}^{\textup{cal}}_U}(R(q)\leq \hat{R}^{+}(q))\geq 1-\delta \ \ \  \forall q,$$
    where $\hat{R}^+(q)=\hat{R}^+_{U}(q)+\hat{R}^{\textup{clip}+}_{\textup{rect}}(q)$.
\end{lemma}
\noindent The proof is detailed in Appendix~\ref{app:proof}.

Armed with a valid UCB $\hat{R}^{+}(q)$, we can rigorously tune the hyper-parameter $\hat{q}$ by applying RCPS as described in  Algorithm~\ref{alg:ss_calib}. This algorithm presents the exact procedure of finding the UCB, which is used to derive $\hat{q}$ that satisfies~\eqref{eq:pac-coverage-new}.

\begin{algorithm}[h]
\caption{Semi-Supervised RCPS for a Binary Loss}\label{alg:ss_calib}
\textbf{Input:} \\
$\mathcal{D}_{L}^{\text{cal}}=\{(X_i,Y_i)\}_{i=1}^{n}$; $\mathcal{D}_{U}^{\text{cal}}=\{X_j\}_{j=n+1}^{n+N}$; \\
$\{\tilde{Y}_j\}_{j=1}^{n+N}$; $\mathcal{Q}_\text{grid}$;
binary loss $L$; \\
control levels $(\alpha, \delta_1, \delta_2)$. \\
\textbf{Process:}\\
        \For {$q$ in $\mathcal{Q}_\textup{grid}$}
        {Compute the unlabeled emp. risk $\hat{R}_{U}(q)$ \\
        Derive the exact UCB $\hat{R}^+_U(q)$ at level $\delta_1$ \\
        Compute the rectifying emp. risk $\hat{R}^{\text{clip}}_{\text{rect}}(q)$ \\
        Derive the exact UCB $\hat{R}^{\text{clip}+}_{\text{rect}}(q)$ at level $\delta_2$ \\
        Derive $\hat{R}^+(q)=\hat{R}^+_U(q)+\hat{R}^{\text{clip}+}_{\text{rect}}(q)$\\
        \eIf{$\hat{R}^+(q)<\alpha$}
        {$\hat{q} \leftarrow q$}
        {break}}

\textbf{Output:} The calibrated parameter $\hat{q}$.
\end{algorithm}

\begin{proposition} 
\label{thm:valid-ss-cal}
Under the assumptions of Proposition~\ref{thm:valid-ss-cal-blocks}, the calibrated parameter $\hat{q}$ constructed by Algorithm~\ref{alg:ss_calib} satisfies~\eqref{eq:pac-coverage-new} for a binary loss function.
\end{proposition}
\noindent The above states that the risk controlling guarantee holds in finite samples, resembling the result of Proposition~\ref{thm:valid-ss-cal-blocks}. Our experiments show that for small sample sizes, the specialized semi-supervised calibration with a binary loss is more powerful than the general method from Section~\ref{sec:ss_RCPS_general}. However, with approximately a few thousand labeled samples, the general semi-supervised RCPS achieves a tighter bound.


As a concluding remark, we discuss the effect of the clipping operator. First, in situations where $L(q) \geq \tilde{L}(q)$ almost surely we have $R_{\text{rect}}^{\text{clip}}(q)=R_{\text{rect}}(q)$. In turn, in this special case, the clipping operator does not add any conservatism. In Section~\ref{sec:image_class}, we provide an example of such a case when constructing prediction sets.
More generally, the clipping operator can introduce a conservative bias for $R^{\text{PP}}$, as shown in \eqref{eq:pp-clip}, however, it reduces the variance of the empirical prediction-powered risk $\hat{R}^{\text{PP}}(q)$. This is attributed to the fact that $\text{Var}((L(q)-\tilde{L}(q))_+) \leq \text{Var}(L(q)-\tilde{L}(q))$.
A rigorous mathematical explanation of the variance reduction is in Appendix~\ref{app:proof}. 
As a result, the clipped version of the rectifying risk $R_{\text{rect}}^{\text{clip}}(q)$ can lead to a tighter UCB for the conservatively biased $R_{\text{PP}}^{\text{clip}}(q)$.

\section{Experiments}

We evaluate the applicability of our methods in two real-data experiments. The first experiment involves constructing prediction sets for an image classification task. The second experiment focuses on classifying movie reviews as favorable or unfavorable in an NLP task, aiming for early termination of the inference process while controlling the trade-off between accuracy and early termination.

\subsection{Few Shot Calibration: Image Classification}
\label{sec:image_class}

We consider a multi-class classification problem using the ImageNet dataset. This dataset consists of pairs of an image $X\in\mathcal{X}$ and its label $Y\in\mathcal{Y} =\{1,\ldots,1000\}$  that indicates the identity of the object in that image. Our goal is to utilize a pre-trained classifier $\hat{f}$ to form a set of plausible labels $\mathcal{T}_{q}(X)$ for a given test instance $X$ that covers the true unknown label $Y$ with high probability. In turn, the risk we want to control is the expectation of the binary miscoverage loss $\mathbb{E}[\mathbbm{1}\{Y \in \mathcal{T}_{q}(X)\}]$.

To formalize the prediction set function, we follow the conformal prediction approach and define $s(X,Y)\in \mathbb{R}$ as a real-valued score function that represents the degree of agreement between the model's prediction $\hat{f}(X)$ and the true label $Y$. This score function can take any form, but we use the convention that a higher score indicates a higher prediction error. With this in place, we define the prediction sets as $\mathcal{T}_q(X)=\{y\in\mathcal{Y}:s(X,y)\leq q\}$, i.e., we include in $\mathcal{T}_q(X)$ all the labels $y$ whose corresponding scores are smaller than or equal to $q$.
In the following experiments, we use the \textit{adaptive prediction sets} (APS) score, introduced in \cite{romano2020classification}, which sums the sorted estimated class probabilities derived by the classifier $\hat{f}(X)$ up to the rank of $Y$. The exact formulation of this score function is given in Appendix~\ref{app:imagenet}.

We compare between five different methods:
\begin{itemize}
    \item \texttt{RCPS lab}---the vanilla RCPS procedure, applied to calibration data that only contains labeled samples. For a fair comparison, the exact Clopper–Pearson bound~\eqref{eq:exact_bound} is used.
    \item \texttt{SS-RCPS CLT}: our semi-supervised RCPS for a general loss from Section~\ref{sec:ss_RCPS_general}, implemented with the CLT UCB. In this case, the risk-controlling guarantee is only asymptotically valid.
    \item \texttt{SS-RCPS WSR}: our semi-supervised RCPS for a general loss with the WSR UCB. In contrast to the CLT approach, the WSR bound is valid in finite samples. We use the WSR method since it has been shown to be more powerful than other non-asymptotic bounds \cite{bates2021distribution}.
    \item \texttt{SS-RCPS BIN}: our specialized semi-supervised RCPS for a binary loss from Section~\ref{sec:ss_RCPS_binary}. Here, we apply the exact Clopper–Pearson bound~\cite{clopper1934use} that requires budget splitting between the labeled and unlabeled risks.
    \item \texttt{RCPS imp}: a naive, invalid RCPS method, applied to a calibration set that includes the hold-out labeled data as well as the imputed unlabeled samples. We refer to this as a naive approach as it does not account for erroneous imputations. We include this method in our experiment to underscore the need for our rigorous semi-supervised methods that leverage unlabeled data while accounting for incorrect imputations to ensure valid results. In our experiments, we employ a slightly modified version of APS for this method, 
    which is even more conservative than the standard APS in terms of coverage, see Appendix~\ref{app:imagenet} for details.
\end{itemize}

To implement the semi-supervised calibration methods, we impute the unknown labels of the unlabeled samples using the most likely, top-1 predictions obtained from a pre-trained Resnet50 classifier, which has a \mbox{top-1} accuracy of approximately 81\%. This classifier is also used to compute the APS scores and construct the prediction sets. Consequently, by design, we have $\tilde{L}(q)\leq L(q)$ almost surely, since $s(X,\tilde{Y})\leq s(X,Y)$ almost surely. This is an example of a situation where the clipping operator does not introduce any bias, as $R_{\text{PP}}(q)=R_{\text{PP}}^{\text{clip}}(q)$ from~\eqref{eq:pp-clip}. 

To create the test set as well as the labeled and unlabeled calibration sets, we randomly split the 50,000 samples from the original ImageNet validation set. Specifically, 25,000 samples are used for testing, and the remaining 25,000 samples are used to form the two disjoint labeled and unlabeled calibration sets. In each experiment, we evaluate the empirical coverage rate---defined as one minus the empirical risk---and the prediction set sizes obtained by each calibration method. We set the error levels to $(\alpha,\delta)=(0.15,0.1)$. For the \texttt{SS-RCPS BIN}, we fix $(\delta_1,\delta_2)=(0.01,0.09)$ when the labeled calibration set is of size less
than 1000, and $(\delta_1,\delta_2)=(0.05,0.05)$ otherwise.

\begin{figure}[t]
    \centering
    \includegraphics[scale=0.32]{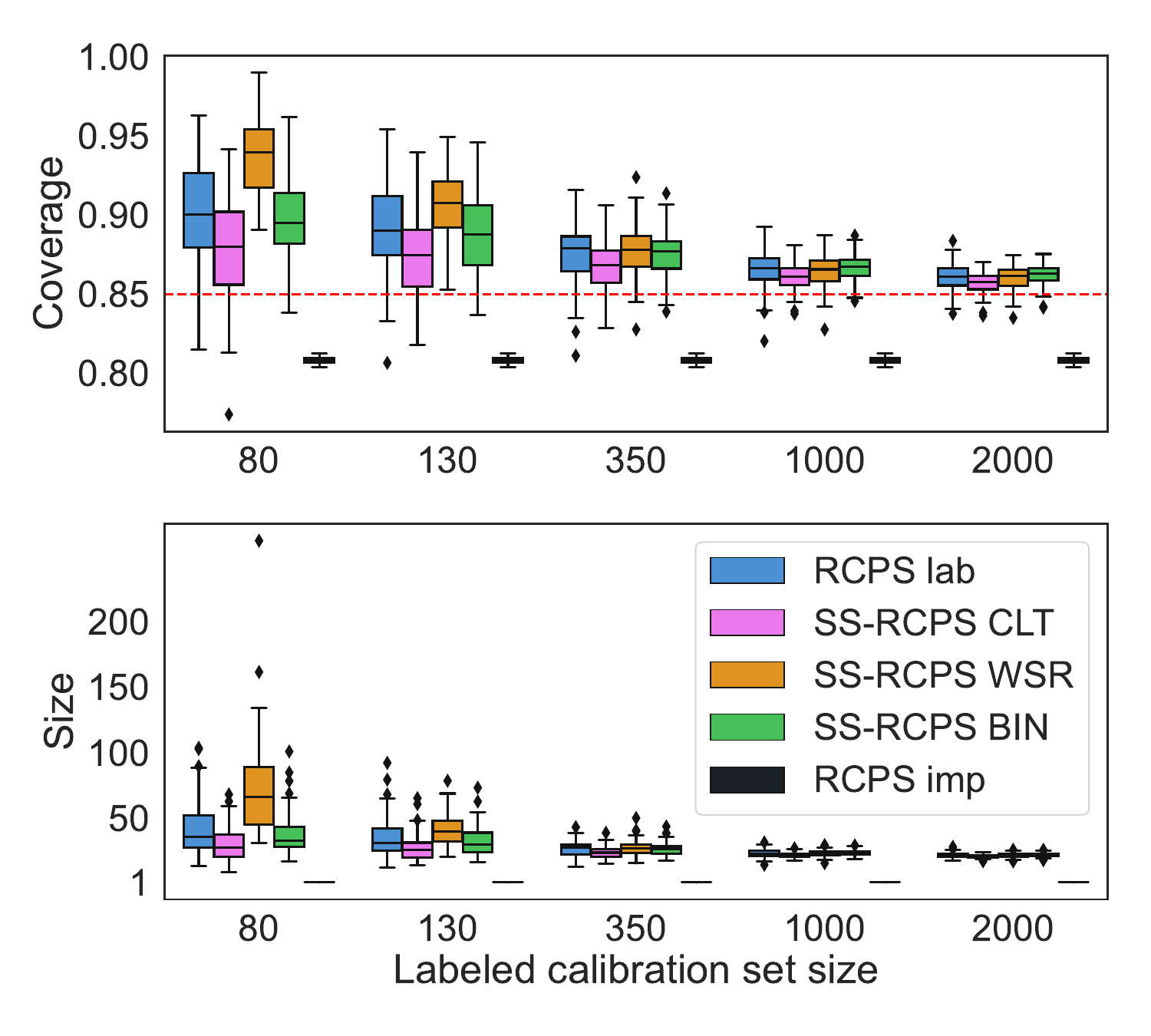}
    \caption{\textbf{ImageNet classification experiment: the effect of labeled calibration set size.} The top panel shows the coverage rate of the constructed prediction sets, while the bottom panel displays the average size of the prediction sets. Both performance metrics are presented as a function of the number of labeled calibration samples and evaluated over 100 different random splits of the data. Refer to the main text for a description of the five different methods compared.}
    \label{Fig:coverage_n_calib}
\end{figure}

Figure~\ref{Fig:coverage_n_calib} displays the empirical distribution of the coverage and set-size metrics as a function of the number of labeled calibration samples. Following the top panel of Figure~\ref{Fig:coverage_n_calib}, we can see that the vanilla \texttt{RCPS lab} controls the risk with high probability, as expected. Observe that when the sample size of the labeled calibration set is small the empirical coverage rate is conservatively controlled and the coverage variance is relatively large. In contrast, the semi-supervised calibration with the CLT bound (\texttt{SS-RCPS CLT}) is less conservative in terms of coverage, but this bound is not always valid. For example, with 130 labeled samples, the coverage rate is below the desired 85\% nominal level 18\% of the time instead of 10\%. 
The finite-sample semi-supervised approach for general loss with the WSR bound (\texttt{SS-RCPS WSR}) is valid across the board, yet, this approach is more conservative than \texttt{SS-RCPS CLT}, especially for small sample sizes. Notably, \texttt{SS-RCPS WSR} tends to surpass both the vanilla \texttt{RCPS lab} and the specialized semi-supervised calibration method (\texttt{SS-RCPS BIN}) as the sample size increases.
The specialized method, however, outperforms both the vanilla \texttt{RCPS lab} and \texttt{SS-RCPS WSR} in the small sample-size regime.
Lastly, the naive \texttt{RCPS imp} method that utilizes all imputed labels without correction leads to an invalid coverage rate. This is anticipated as $\tilde{L}(q)\leq L(q) \Rightarrow R_U(q)\leq R(q)$ in our experiments. In turn, by naively augmenting the small labeled data with the large imputed data we control a risk that is smaller than the desired $R(q)$. Consequently, the naive calibrated parameter $\hat{q}$ is too small, failing to achieve the desired coverage rate.
Finally, the bottom panel of Figure~\ref{Fig:coverage_n_calib} portrays the average size of the prediction sets.
As expected, methods that achieve lower coverage levels result in smaller prediction sets.

\begin{figure}[ht]
    \centering
    \includegraphics[scale=0.395]{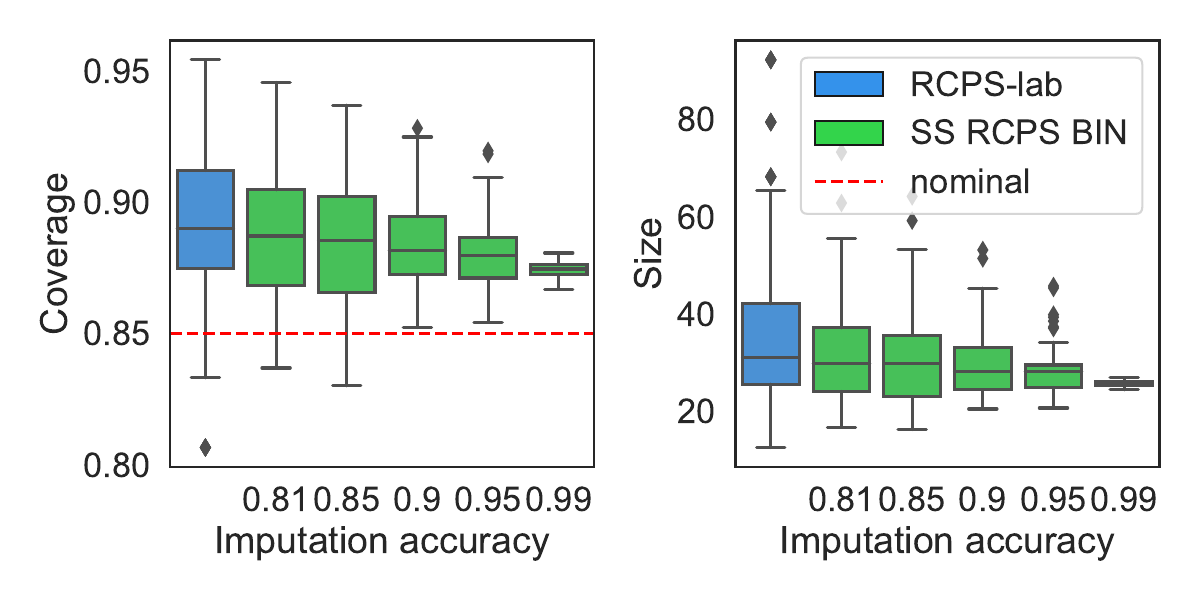}
    \caption{\textbf{ImageNet classification experiment: the effect of the imputation accuracy.} The performance metrics are evaluated as a function of the accuracy of the imputed labels. All other details are as in Figure~\ref{Fig:coverage_n_calib}.}
    \label{Fig:coverage_accuracy}
\end{figure}

In the next experiment, we examine how the accuracy of imputed labels affects the performance of our semi-supervised calibration method. We fix the size of the labeled calibration set at 130 and synthetically increase the imputation accuracy by gradually replacing the incorrect imputed labels obtained by the classifier with the true labels. We focus on the \texttt{SS-RCPS BIN} method, which---as shown in Figure~\ref{Fig:coverage_n_calib}---performs best in this challenging small sample-size regime. Similarly to the experiment in Figure~\ref{Fig:coverage_n_calib}, we construct the prediction sets using the same pre-trained Resnet classifier. The results, summarized in Figure~\ref{Fig:coverage_accuracy}, show that as imputation accuracy improves, the coverage variance decreases and the gap between the average coverage and the nominal level narrows. In line with this trend, the average size of prediction sets decreases as the accuracy of the imputed labels increases. Further, we can see that the average coverage remains somewhat conservative even with accurate imputations due to the budget splitting of the specialized method. 
Overall, this experiment highlights that significant improvement in calibration performance can be obtained when relying on accurate imputed labels, as expected. 
Importantly, the risk-controlling guarantee 
holds, regardless of the imputation accuracy.




\subsection{Early Time Series Classification}
\label{sec:exp_nlp}

The goal of early time series classification (ETSC) is to predict the label of a given input data stream as quickly as possible. Specifically, we consider an NLP application using the Internet-Movie-Database (IMDb)~\cite{maas-EtAl:2011:ACL-HLT2011} that includes movie reviews labeled as positive or negative. We use an LLM to analyze a given review and predict its label. Naturally, the inference time increases with the number of tokens processed, making it desirable to terminate the inference as soon as the necessary information for making the prediction is obtained. In this context, we follow the risk-controlling method presented in~\cite{ringel2024early} and seek an algorithm that ensures the gap in accuracy between the full- and partial-processing of the text is controlled.

More formally, denote by $X = (X^1, X^2, \ldots, X^{t_{\text{max}}}) \in \mathcal{X}$ an observed input sequence of tokens with a maximum length of $t_{\text{max}}$. Denote further by $Y \in \mathcal{Y} = \{0,1\}$ the corresponding label, indicating whether the review $X$ is positive or negative. We assume access to a pre-trained model $\hat{f}$ that processes the input $X$ sequentially and, at each timestep $t$, predicts the label given $X^{\leq t} = (X^1, \ldots, X^t)$.

We employ a stopping rule function that decides whether to stop the inference process at each timestep $t$. The idea is to halt the inference once the model is ``confident enough'' in its prediction based on the data observed up to that point. In more detail, let  $\hat{\pi}(X^{\leq t})$ be some heuristic notion of the classifier's confidence in its prediction based on $X^{\leq t}$. For example, $\hat{\pi}(X^{\leq t})$ can be the largest softmax value of a neural net classifier. Consequently, we halt the inference the first time $\hat{\pi}(X^{\leq t})$ passes a time-dependent threshold $\underline{\hat{q}}_t \in [0,1]$. Formally, the stopping time, denoted by $\tau_{\hat{\underline{q}}}(X)$, is formulated as
\begin{equation}
\tau_{\hat{\underline{q}}}(X) = \min \{t:\hat{\pi}(X^{\leq t})\geq \underline{\hat{q}}_t \ \text{or} \ t=t_\text{max}\}, 
\end{equation}
where $\underline{\hat{q}} = (\underline{\hat{q}}_1, \underline{\hat{q}}_2,\ldots,\underline{\hat{q}}_{t_{\text{max}}})$ is the hyper-parameter vector of thresholds.




The task here is to find such a hyper-parameter vector $\underline{q} \in[0,1]^{t_{\text{max}}}$ that controls the accuracy gap conditioned on the halt time being less than or equal to $t$~\cite{ringel2024early}. In turn, the risk is the proportion of samples for which the classifier's prediction is correct when applied to the entire sequence but incorrect when applied only up to the early timestep $\tau_{\hat{\underline{q}}}(X)$. Formally, let $\hat{Y}_{\text{early}}(\hat{\tau})$ and $\hat{Y}_{\text{full}}$ be the predicted labels obtained by $X^{\leq \tau_{\hat{\underline{q}}}(X)}$ and the full sequence $X$, respectively. The risk is then defined as
\begin{equation}
\label{eq:R_leq_t}
    R_{\text{gap}}^{\leq t}(\hat{\underline{q}}) = \mathbb{E} \left [ L_{\text{gap}}(Y,\hat{Y}^{\text{full}},\hat{Y}^{\text{early}}(\tau_{\hat{\underline{q}}})) \mid \tau_{\hat{\underline{q}}}(X) \leq t \right ],
\end{equation}
where
\begin{align}
  \label{eq:accuracy_gap}
  &L_{\text{gap}}(Y,\hat{Y}^{\text{full}},\hat{Y}^{\text{early}}(\tau_{\hat{\underline{q}}})) = \\
  & \quad \quad \quad \left(\mathbbm{1}\{Y = \hat{Y}^{\text{full}}\} - \mathbbm{1}\{Y = \hat{Y}^{\text{early}}(\tau_{\hat{\underline{q}}})\} \right)_+.
\end{align}
Notice that the loss $L_{\text{gap}}$ is binary due to the clipping operator. In~\cite{ringel2024early}, the authors present an algorithm that rigorously attains risk control conditional on the accumulated halt times:
\begin{equation}
\label{eq:conditional_guarantee}
\begin{aligned}
\mathbb{P}_{\mathcal{D}_L^{\text{cal}}} \left ( R_{\text{gap}}^{\leq t}(\hat{\underline{q}}) \leq \alpha \ \text{for all} \ t\geq t_0 \right ) \geq 1-\delta,
\end{aligned}
\end{equation}
where $t_0$ is defined as the first timestep for which $\mathbb{P}( \tau_{\hat{\underline{q}}}(X) \leq t_0) > 0$, as otherwise \eqref{eq:R_leq_t} is undefined.
This conditional risk control guarantee prevents the prediction rule from performing poorly on sequences with early halt times, leading to a more reliable system.

Since the risk~\eqref{eq:R_leq_t} is not monotonous with respect to the hyper-parameter $\hat{\underline{q}}$, the authors of~\cite{ringel2024early} suggested using the \textit{learn then test} framework~\cite{angelopoulos2025learn}. Nonetheless, as we argued in Section~\ref{sec:warm_up}, RCPS can also control non-monotonic risks, which allows us to apply it here as well. 
To define $\mathcal{Q}_{\text{grid}}$, the authors suggest splitting the labeled calibration set into two approximately equal disjoint sets, using the first set for $\mathcal{Q}_{\text{grid}}$ definition and the second for calibration. A key challenge of the calibration process is that when aiming to achieve conditional risk control, the problem of small sample sizes is inevitable, leading to overly conservative UCBs and preventing early halt times. This greatly motivates the application of our semi-supervised calibration methods, which can effectively increase the calibration set size without invalidating~\eqref{eq:conditional_guarantee}.
The only difference is that, in our setup, the probability in~\eqref{eq:conditional_guarantee} is over both labeled and unlabeled calibration sets. In the interest of space, the complete calibration process for tuning $\underline{\hat{q}}$ can be found in Appendix~\ref{app:NLP}.

In our experiment, the Phi-3-Mini-4K-Instruct LLM~\cite{abdin2024phi3technicalreporthighly} is used to impute the unknown labels of the unlabeled calibration set, and the same model is used for calibration and evaluation. This model is accessible via \href{https://huggingface.co/microsoft/Phi-3-mini-4k-instruct}{HuggingFace} and has a top-1 accuracy of approximately 93\%. Our datasets include a few labeled calibration samples, 50,000 unlabeled calibration samples, and 16,000 test samples. We compare all methods presented in Section~\ref{sec:image_class} except the invalid \texttt{RCPS imp} method. We set the error levels of~\eqref{eq:conditional_guarantee} to be $(\alpha,\delta)=(0.1,0.01)$. For the \texttt{SS-RCPS BIN}, we fix $(\delta_1,\delta_2)=(0.001,0.009)$.

\begin{figure*}[t]
    \centering
    \subfloat[]
    {\includegraphics[scale=0.308]{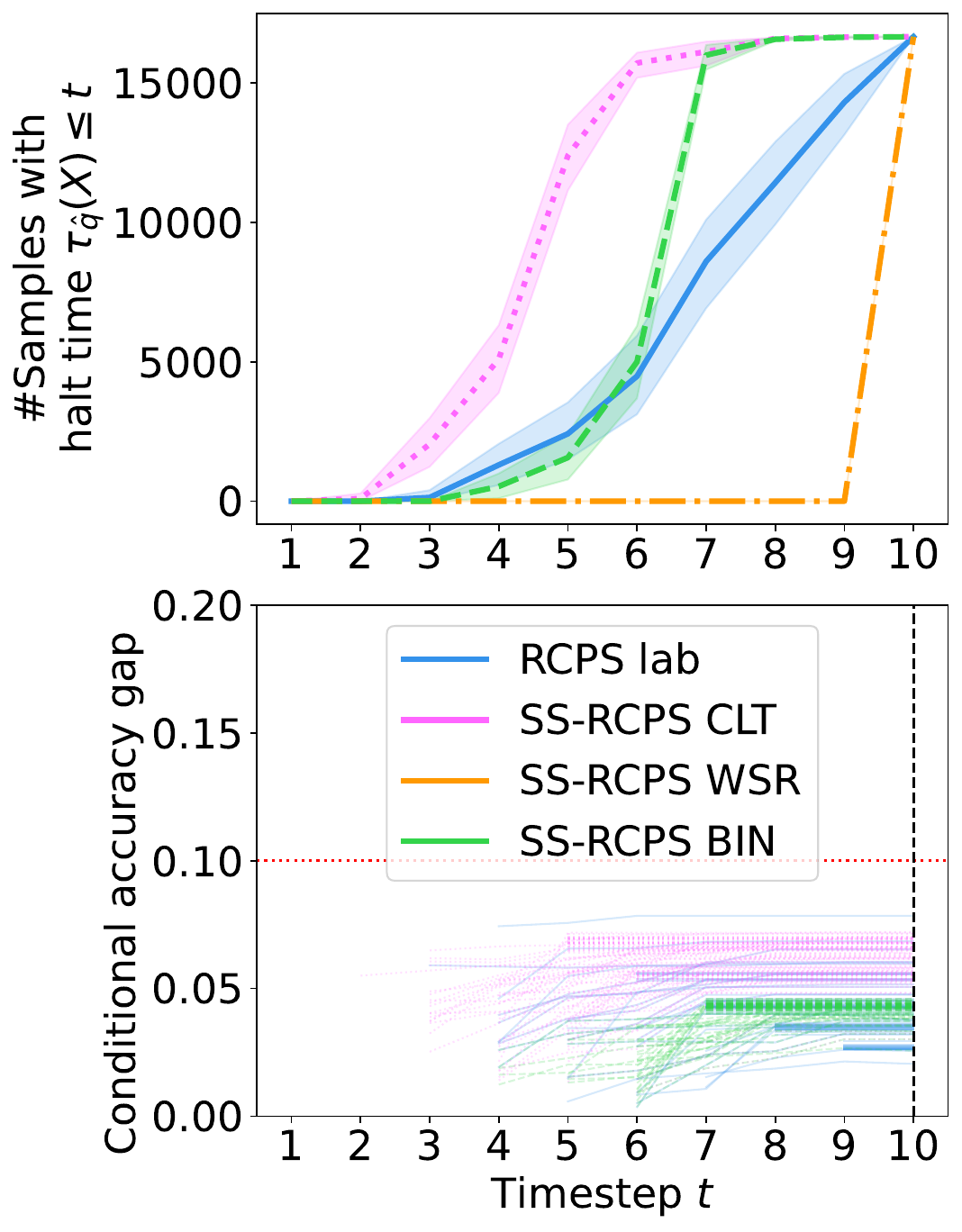}}
    \subfloat[]
    {\includegraphics[scale=0.308]{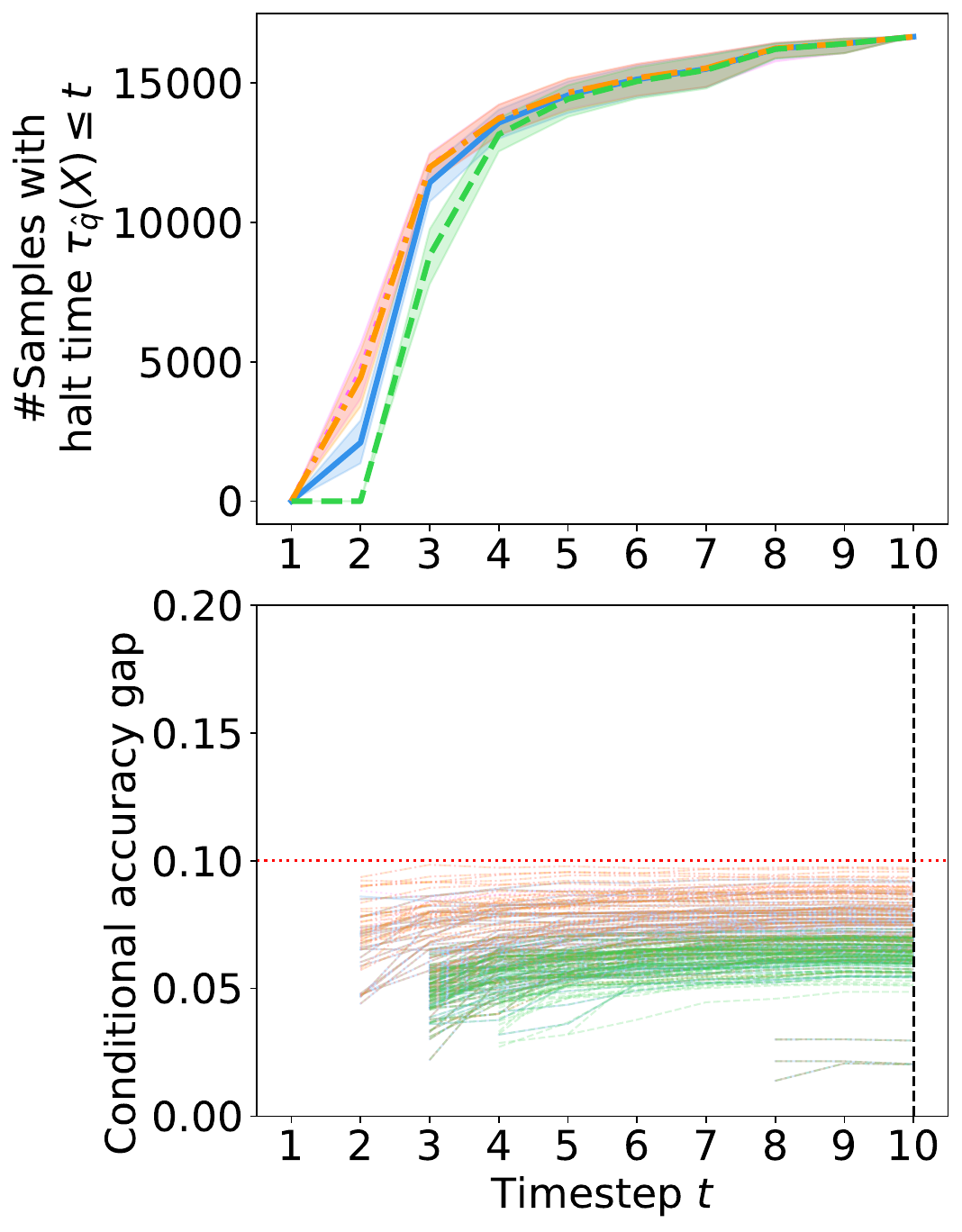}}
    \caption{\textbf{ETSC experiment on NLP task: the effect of the size of the labeled calibration set.} The vanilla and semi-supervised RCPS methods implemented with a labeled calibration set of size (a) 300 samples, and (b) 4000 samples. Top panels present the accumulated halt times as a function of $t$, averaged over 100 random data splits; the shaded area represents a 95\% confidence interval. Bottom panels present the corresponding empirical conditional accuracy gap, $\hat{R}_{\text{gap}}^{\leq t}(\hat{\underline{q}})$, across the 100 trials. Each curve corresponds to a different random split of the labeled calibration and test data. }
    \label{Fig:imdb_300}
\end{figure*}

Figure~\ref{Fig:imdb_300} (a) illustrates the accumulated halt times (top) and empirical conditional accuracy gap (bottom), when using 300 labeled calibration samples. Following the top panel of that figure, the \texttt{SS-RCPS CLT} achieves the earliest halt times, but this method is not guaranteed to be valid in finite samples. The \texttt{SS-RCPS WSR} does not stop early at all when the sample size is small, which goes in hand with the ImageNet experiment. The \texttt{SS-RCPS BIN} method outperforms \texttt{RCPS lab} for most time steps, both in terms of earlier halt times and variance reduction. Figure~\ref{Fig:imdb_300} (b) repeats the same experiment except that we use a larger labeled calibration set of size 4000 samples. Here, the \texttt{SS-RCPS BIN} method loses its advantage over the other methods, while \texttt{SS-RCPS WSR} outperforms \texttt{RCPS lab} and performs comparably to \texttt{SS-RCPS CLT}. 
This conclusion is consistent with the ImageNet experiment, where \texttt{SS-RCPS WSR} demonstrated better performance with larger labeled calibration data. 
Notably, all methods control the conditional accuracy gap, indicated by the bottom panels of Figures~\ref{Fig:imdb_300}.

\section{Conclusions}
We presented a novel approach for effectively leveraging unlabeled data to improve the statistical efficiency of RCPS. The key properties of our methods are as follows: (1) when the imputation model is relatively accurate, our semi-supervised calibration procedures successfully reduce the conservatism and variance of the prediction rule's error rate; and (2) all the presented methods provably remain valid even when handed inaccurate imputations. With that said, when the imputations are inaccurate our approach may not improve the statistical efficiency of the vanilla RCPS which relies only on labeled samples. This calls for the design of imputation strategies and score functions to mitigate this issue, as an alternative to using the most likely predicted label. Another limitation is the i.i.d. assumption required to ensure the validity of our methods. 
A promising future direction would be to extend our approach to handle covariate (or label) shift, possibly by a reweighting mechanism of the loss functions~\cite{angelopoulos2023prediction,parkpac,tibshirani2019conformal}. 
Another promising direction is to derive a p-value or UCB for a trinomial variable, which is the actual loss we aim to control, 
thereby eliminating the need for the conservative clipping operator.

\section*{Acknowledgments}
This research was supported by the European Union (ERC, SafetyBounds, 101163414). Views and opinions expressed are however those of the authors only and do not necessarily reflect those of the European Union or the European Research Council Executive Agency. Neither the European Union nor the granting authority can be held responsible for them. This research was also partially supported by the Israel Science Foundation (ISF grant 729/21). Y.~R. acknowledges additional support from the Career Advancement Fellowship at the Technion. The authors are deeply grateful to Anastasios N. Angelopoulos for his insightful comments and valuable feedback.

\bibliographystyle{unsrt}
\bibliography{biblio}

\clearpage
\begin{appendices}
\begin{appendices}

\section{Additional Algorithmic Details and Proofs}
\subsection{Fixed Sequence Testing}
\label{app:fst}
We begin by presenting more details regarding RCPS when the risk is not monotonous with respect to the hyper-parameter $q$. In this case, the risk-control task is reformulated as a multiple-hypothesis testing task.
We pre-specify a grid search for the parameter $\hat{q}$: $\mathcal{Q}_\text{grid}=\{q_1,q_2,\ldots, q_N\}$ and compute a UCB with error level $\delta$ at these specifies points $(\hat{R}^+(q_1),\hat{R}^+(q_2),\ldots, \hat{R}^+(q_N))$. For each value, we define the null hypothesis: $H_{0,i}: R(q_i)>\alpha,$ indicating that the risk is not controlled. We iterate over the grid and set $\hat{q}$ to be $q_i$ if $\hat{R}(q_i)\leq \alpha$. We stop the iteration at the first instance where we cannot reject the null hypothesis. This procedure is called fixed-sequence-testing and is depicted in the following algorithm.


\begin{algorithm}[h]
\caption{Fixed Sequence Testing}\label{alg:fst}
\textbf{Input:} risk control level $\alpha \in(0,1)$, parameter grid $\mathcal{Q}_\text{grid}=\{q_1,q_2,\ldots, q_N\}$ and an upper confidence bound with error level $\delta$ at these specifies points $(\hat{R}^+(q_1),\hat{R}^+(q_1),\ldots, \hat{R}^+(q_N))$.

\textbf{Process:}\\
        \For {$q$ in $\mathcal{Q}_\textup{grid}$}
        {\eIf{$\hat{R}^+(q)<\alpha$}
        {$\hat{q} \leftarrow q$}
        {break}}

\textbf{Output:} $\hat{q}$.
\end{algorithm}

Next, we will discuss the validity of this procedure. Let $V$ be the number of true nulls that are falsely rejected by the testing procedure, and define the \textit{family-wise-error-rate} (FWER) as the probability of falsely rejecting at least one true null hypothesis, i.e., $\text{FWER}=\mathbb{P}(V\geq1)$. In multiple hypothesis testing, our goal is to design a testing procedure that ensures the FWER does not exceed $\delta$, meaning $\mathbb{P}(V\geq1)\leq \delta$. Theorem~\ref{thm:fst} states that RCPS with fixed-sequence-testing controls the FWER. Intuitively the FWER is controlled because the first true null encountered is not rejected with a probability of at least $1-\delta$, ending the procedure. Notice that this stopping rule is precisely used in all the algorithms we present in the main text; see Algorithms~\ref{alg:ss_calib_blocks} and \ref{alg:ss_calib} of the main manuscript.
The proof for the validity of RCPS with fixed sequence testing is provided below, it is a direct consequence of \cite{angelopoulos2025learn}.

\begin{proof}[Proof of Theorem~\ref{thm:fst}]
    Denote the hypothesis $H_{0,i}: R(q_i)>\alpha$, meaning the risk is not controlled. Denote by $H_{0,j}$ the $j$-th ordered hypothesis. We reject the null if $\hat{R}^+(q_i)\leq\alpha$ and otherwise accept it. If all the hypotheses are false, then $P(V\geq 1)=0$ holds trivially. 
    Let $j_0$ denote the index of the first true null hypothesis, i.e., $H_{0,j_0}$ is true while all the preceding $H_{0,j'}$ for $j'<j_0$ are false. Due to the construction of the fixed sequence testing procedure, we will encounter this first true null hypothesis only at step $j_0$.
    Now, observe that 
    \begin{align}
        &P(V\geq 1)=1-P(V=0)\\
        &=1-P(\hat{R}^+(q_{j_0})>\alpha\mid R(q_{j_0})>\alpha)\\
        &= P(\hat{R}^+(q_{j_0})\leq\alpha\mid R(q_{j_0})>\alpha)\leq \delta.
    \end{align}
    Above, the second equality holds since the testing procedure halts the first time that the UCB exceeds $\alpha$, and thus we get $V=0$ if and only if $\hat{R}^+(q_{j_0})>\alpha$;
    under this event, the procedure would terminate without rejecting $H_{0,j_0}$ and $H_{0,j'}, \forall j'>j_0$.
    The last inequality follows from the validity of the upper confidence bound. 
\end{proof}

\subsection{Waudby-Smith-Ramdas UCB}
\label{app:wsr}
For completeness, we provide here the algorithm for constructing the WSR UCB. Following~\cite{jazbecfast}, we slightly modify the original algorithm from \cite{waudby2024estimating}, to account for random variables bounded between $A$ and $B$. The Proposition below states the validity of the modification we introduced. This statement is a direct consequence of the validity of the original WSR UCB~\cite{waudby2024estimating}.
\begin{proposition}
    Under the i.i.d. assumption, the WSR-based UCB $\hat{R}^+_{\text{WSR}}$ constructed by Algorithm~\ref{alg:wsr} for random variables bounded between $A$ and $B$ satisfies~\eqref{eq:bound}.
\end{proposition}

\begin{algorithm}
\caption{Waudby-Smith-Ramdas UCB}\label{alg:wsr}
\textbf{Input:} \\ $n$ i.i.d. bounded random variables $W_i\in[A,B]$; \\ error level $\delta$.

\textbf{Process:}\\
        \For {$i=1,\ldots,n$}
        {$\hat{\mu}_i=\frac{\frac{1}{2}+\sum_{j=1}^iW_j}{1+i}$,\\
        $\hat{\sigma}^2_i=\frac{\frac{1}{4}+\sum_{j=1}^i(W_j-\hat{\mu}_j)^2}{1+i}$,\\
        $\nu_i=\min\left\{\frac{1}{B-A},\sqrt{\frac{2\log(1/\delta)}{n\hat{\sigma}^2_{i-1}}} \right\}$,\\
        $\mathcal{K}_i(R)=\prod_{j=1}^{i}\{1-\nu_j(W_j-R)\}$.\\
        }
        
\textbf{Output:} A finite-sample valid UCB $$\hat{R}^+_{\text{WSR}}=\inf\left\{R\geq 0: \max_{i=1,\ldots,n}\mathcal{K}_i(R)>\frac{1}{\delta} \right\}.$$
\end{algorithm}

\begin{proof}
    First, if $W_j\in[A,B]$ then: $$(W_j-\mathbb{E}(W_j))\in[A-B,B-A].$$
    Further, notice that: $$\nu_i=\min\left\{\frac{1}{B-A},\sqrt{\frac{2\log(1/\delta)}{n\hat{\sigma}^2_{i-1}}} \right\}\in[0,\frac{1}{B-A}]$$ since the term $\sqrt{\frac{2\log(1/\delta)}{n\hat{\sigma}^2_{i-1}}}$ is always non-negative. Therefore, $$\nu_i(W_j-\mathbb{E}(W_j))\in[-1,1],$$ and $$1-\nu_i(W_j-\mathbb{E}(W_j)\geq 0.$$ Finally, this implies that $$\mathcal{K}_i=\prod_{j=1}^{i}\{1-\nu_j(W_j-\mathbb{E}(W_j))\}$$ is a non-negative martingale. Given this property of $\mathcal{K}_i$, the rest of the proof directly follows from~\cite{waudby2024estimating}.
\end{proof}

Note that linearly scaling $W_i$ to the $[0, 1]$ range and applying the standard WSR UCB is also possible. Since $W_i \in [A,B]$, we first define $V_i = (W_i-A)/(B-A)$, and then compute the standard WSR UCB using $\{V_i\}_{i=1}^n$. Finally, we multiply this UCB by $(B-A)$ and add $A$. Following Figure~\ref{Fig:wsr_comparison}, we can see a little difference in power between the linearly scaled WSR and generalized WSR that extends beyond the standard $[0,1]$ loss.

\begin{figure}[h]
    \centering
    \includegraphics[scale=0.38]{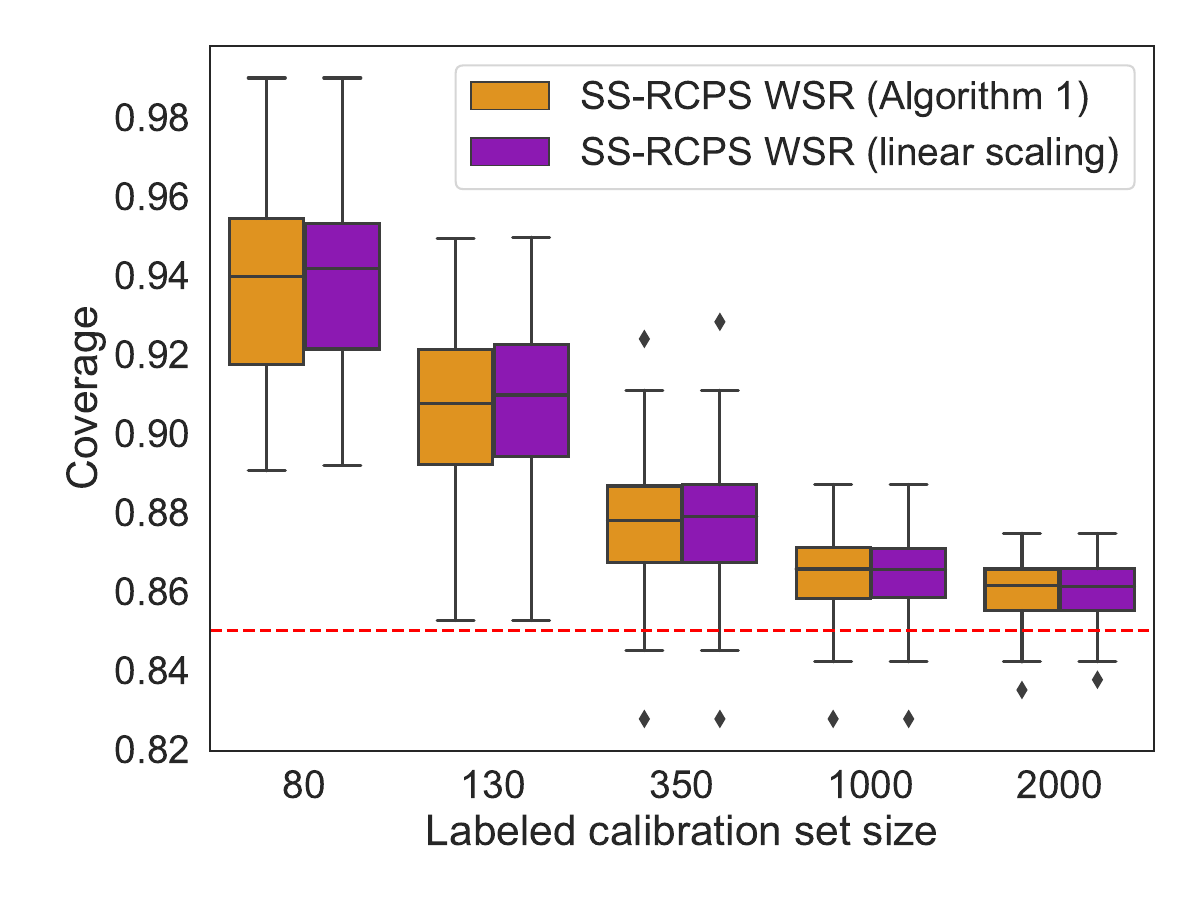}
    \caption{ImageNet classification experiment: the coverage rate of the constructed prediction sets as a function of the number of labeled calibration samples, evaluated over 100 different random splits of the data. Comparison between the modified WSR algorithm for bounded random variables and the original WSR using linear scaling. Other details are as in Figure~\ref{Fig:coverage_n_calib}.}
    \label{Fig:wsr_comparison}
\end{figure}




\section{Mathematical Proofs}
\subsection{Lemma~\ref{thm:valid_bound}}
\label{app:proof}

Here we provide the proof of Lemma~\ref{thm:valid_bound}.

\begin{proof}[Proof of Lemma~\ref{thm:valid_bound}]
Recall that
\begin{align}
     R(q) &= \mathbb{E}[L(q)],\\ 
     R_{U}(q) &=\mathbb{E}[\tilde{L}(q)],\\ 
     R_{\textup{rect}}(q) &=\mathbb{E}[L(q)-\tilde{L}(q)],\\ 
     R^{\textup{clip}}_{\textup{rect}}(q) &=\mathbb{E}[(L(q)-\tilde{L}(q))_+].
\end{align}
Recall also that $$R(q)\leq R_{U}(q)+R^{\textup{clip}}_{\textup{rect}}(q),$$
following~\eqref{eq:unbiased} and~\eqref{eq:pp-clip}.

Therefore, 
\begin{align}
    &\{R_{U}(q)\leq \hat{R}^+_{U}(q)\} \cap \{R^{\textup{clip}}_{\textup{rect}}(q)\leq \hat{R}^{\textup{clip}+}_{\textup{rect}}(q)\} \Longrightarrow\\
    &R(q)\leq\hat{R}^+(q), 
    \label{eq:relation}
\end{align}
where $$\hat{R}^+(q)=\hat{R}^+_{U}(q)+\hat{R}^{\textup{clip}+}_{\textup{rect}}(q).$$

As a consequence of relation~\eqref{eq:relation} we get
\begin{align*}
    &\mathbb{P}(R(q)\leq\hat{R}^+(q)) \\&\geq \mathbb{P}(\{R_{U}(q)\leq \hat{R}^+_{U}(q)\} \cap \{R^{\textup{clip}}_{\textup{rect}}(q)\leq \hat{R}^{\textup{clip}+}_{\textup{rect}}(q)\})\\
    &=1-\mathbb{P}(\{R_{U}(q)> \hat{R}^+_{U}(q)\} \cup \{R^{\textup{clip}}_{\textup{rect}}(q)> \hat{R}^{\textup{clip}+}_{\textup{rect}}(q)\})\\
    &\geq 1- \mathbb{P}(R_{U}(q)> \hat{R}^+_{U}(q))- \mathbb{P}(R^{\textup{clip}}_{\textup{rect}}(q)> \hat{R}^{\textup{clip}+}_{\textup{rect}}(q))\\
    &\geq 1-\delta_1-\delta_2=1-\delta.
\end{align*}

The second inequality holds by the union bound and the last inequality holds by the validity of the UCBs $\hat{R}^+_{U}(q)$ and $\hat{R}^{\textup{clip}+}_{\textup{rect}}(q)$.

\end{proof}

\subsection{Variance Reduction}
Denote by $L,\tilde{L}$ two random variables that take values in $\{0,1\}$. Further denote by $p_1, p_2$ the probabilities: $\mathbb{P}(L=1,\tilde{L}=0)$ and $\mathbb{P}(L=0, \tilde{L}=1)$, respectively.
In what follows, we show that $$\text{Var}(L-\tilde{L}) \geq \text{Var}((L-\tilde{L})_+).$$

We start by calculating $\text{Var}(L-\tilde{L})$, i.e.,
\begin{equation}
\label{eq:var-non-clipped}
    \text{Var}(L-\tilde{L})=\mathbb{E}((L-\tilde{L})^2)-(\mathbb{E}(L-\tilde{L}))^2.
\end{equation}
Utilizing the fact that the losses are binary, the right-hand side in \eqref{eq:var-non-clipped} can be expressed as
\begin{align*}
    \mathbb{E}(L-\tilde{L})=&\mathbb{P}(L=0,\tilde{L}=0)(0-0)\\
    &+\mathbb{P}(L=1,\tilde{L}=0)(1-0)\\
    &+\mathbb{P}(L=0,\tilde{L}=1)(0-1)\\
    &+\mathbb{P}(L=1,\tilde{L}=1)(1-1)\\
    &=p_1-p_2.
\end{align*}
Similarly, the left-hand-side in \eqref{eq:var-non-clipped} can be simplified as follows
\begin{align*}
    \mathbb{E}((L-\tilde{L})^2)=&\mathbb{P}(L=0,\tilde{L}=0)(0-0)^2\\
    &+\mathbb{P}(L=1,\tilde{L}=0)(1-0)^2\\
    &+\mathbb{P}(L=0,\tilde{L}=1)(0-1)^2\\
    &+\mathbb{P}(L=1,\tilde{L}=1)(1-1)^2\\
    &=p_1+p_2.
\end{align*}
Combining the two, we get:
\begin{equation}
    \text{Var}(L-\tilde{L})=p_1+p_2-(p_1-p_2)^2
\end{equation}

We now turn to calculate the variance of the clipped random variable $(L-\tilde{L})_+$, formulated as
\begin{align*}
    \text{Var}((L-\tilde{L})_+)&=\mathbb{P}(L>\tilde{L})-(\mathbb{P}(L>\tilde{L}))^2\\
    &=\mathbb{P}(L=1,\tilde{L}=0)\\
    &-(\mathbb{P}(L=1,\tilde{L}=0))^2\\
    &=p_1-p_1^2.
\end{align*}

Finally, since $p_1-p_1^2\leq p_1+p_2-(p_1-p_2)^2$ we get that $\text{Var}((L-\tilde{L})_+) \leq \text{Var}(L-\tilde{L})$, and equality holds when $p_2=\mathbb{P}(L=0, \tilde{L}=1)=0$, meaning $L\geq \tilde{L}$ almost surely.

\section{Additional Experimental Details}
\subsection{Classification Experiment}
\subsubsection{Score Definition}
\label{app:imagenet}

Here we provide the exact formulation of the APS scores we use in the experiments from Section~\ref{sec:image_class}. The score is given by
\begin{align*}
    &s^{\rm APS}\left(x,y\right)= \\ & \quad \sum_{y'\in\mathcal{Y}}\hat{{\pi}}_{y'}\left(x\right)\mathbbm{1}\left\{\hat{{\pi}}_{y'}\left(x\right)>\hat{{\pi}}_{y}\left(x\right)\right\}
    +\hat{{\pi}}_{y}\left(x\right)\cdot U,
\end{align*}
where $\hat{{\pi}}_{y}\left(x\right)$ is the estimated conditional probability $\mathbb{P}(Y=y\mid X=x)$ obtained by the classifier, and $U\sim \mathrm{Unif}(0,1)$ is a uniform random variable. When evaluating \texttt{RCPS imp}, the vanilla RCPS method that uses all imputed labels without corrections, we set $U$ to 0. 
In general, this random variable allows for empty prediction sets in some cases to avoid overly conservative coverage rates.
When $U$ is set to 0, the prediction sets always include at least the label with the highest predicted probability.
Thus, this modification only increases the conservatism of the average coverage achieved. As explained in Section~\ref{sec:image_class}, we use this modification as a way to increase the coverage rate of \texttt{RCPS imp}, which is invalid even with this modification.

\subsubsection{Power Tuning}
\label{app:baselines_lambda}
In this section, we provide further details on the power parameter $\lambda$ from~\cite{angelopoulos2023ppi++} (introduced in Section~\ref{sec:ss_RCPS_general}) and two variants of our method that incorporate it to attain more controlled variance reduction.

In more detail, the authors of~\cite{angelopoulos2023ppi++} suggest a variant of the prediction-powered risk from~\eqref{eq:PP-risk} by introducing a hyperparameter $\lambda$ that controls the influence of the unlabeled data:

\begin{align}
   \label{eq:PP-risk_efficient}
    &\hat{R}_\text{PP}^{++}(q)= \\
    &\lambda \frac{1}{N}\sum_{j=n+1}^{n+N} \tilde{L}_j(q)+\frac{1}{n}\sum_{i=1}^n L_i(q)-\lambda \frac{1}{n}\sum_{i=1}^n \tilde{L}_i(q).
\end{align}
Notice that the above risk remains unbiased for a fixed $\lambda$. The optimal parameter $\lambda^*$ is then obtained by minimizing the variance of this risk, yielding:

$$\lambda^*=\frac{\text{Cov}(L_i(q),\tilde{L}_i(q))}{(1+\frac{n}{N})\text{Var}(\tilde{L}_i(q))}.$$
Since $\text{Cov}(L_i(q),\tilde{L}_i(q))$ and $\text{Var}(\tilde{L}_i(q))$ are unknown, these must be estimated from the data. As a result, the parameter $\lambda^*$ aims to tighten the semi-supervised UCBs, ensuring they outperform or match those computed with labeled data alone. Notably, since $\text{Cov}(L_i(q),\tilde{L}_i(q))$ and $\text{Var}(\tilde{L}_i(q))$  are estimated on the same holdout data used to construct the UCB, this approach is valid only asymptotically~\cite{angelopoulos2023ppi++}. Conversely, the construction of a finite-sample valid UCB requires additional hold-out data---or splitting the limited available calibration data---to estimate $\lambda^*$.

\begin{figure}[h]
    \centering
    \includegraphics[scale=0.38]{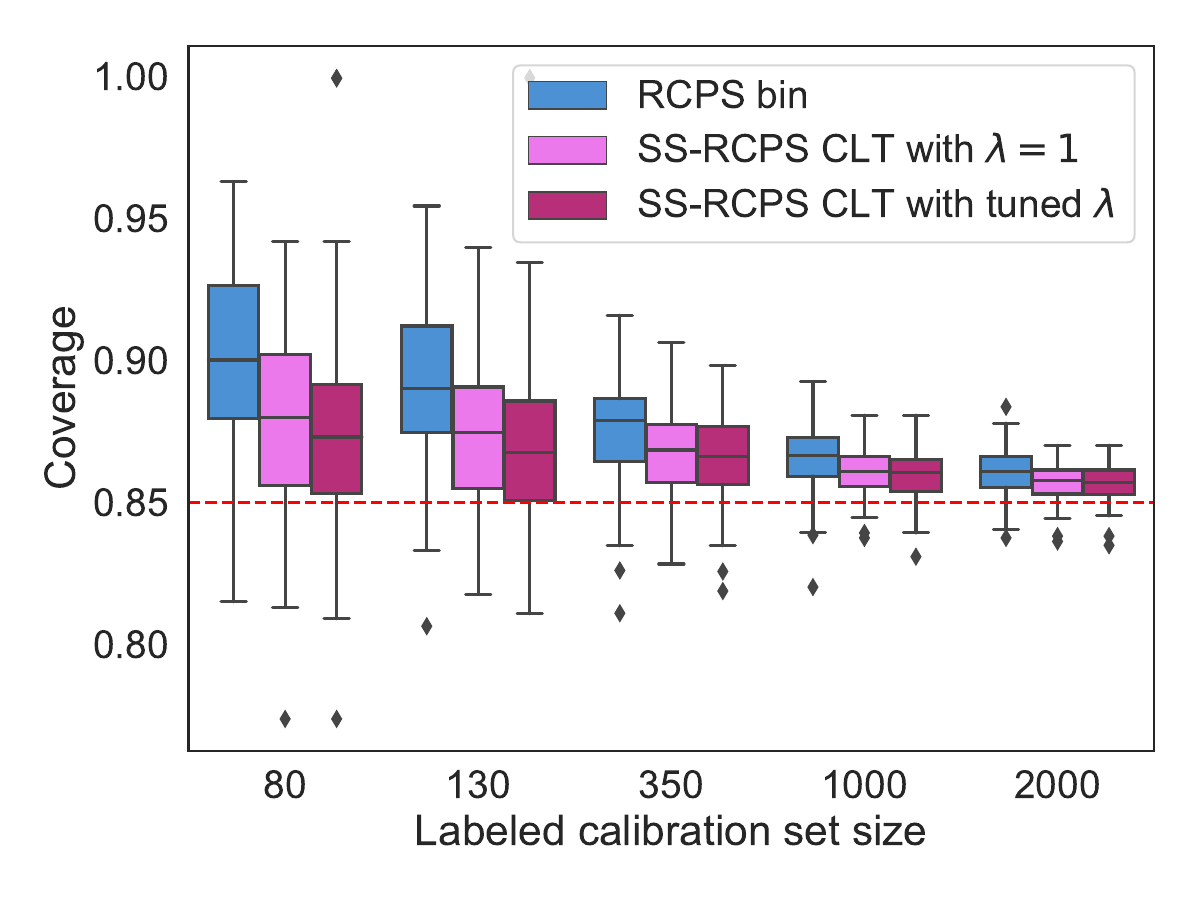}
    \caption{ImageNet classification experiment: the coverage rate of the constructed prediction sets, as a function of the number of labeled calibration samples evaluated over 100 different random splits of the data. Comparison between RCPS using only labeled data, the CLT UCB with $\lambda=1$, and the CLT UCB with the tuned $\lambda$. Other details are as in Figure~\ref{Fig:coverage_n_calib}.}
    \label{Fig:clt_comparison}
\end{figure}

We now turn to introduce two additional variants of our method that utilize $\lambda$ for power tuning. The first computes the UCB using the CLT, as in Algorithm~\ref{alg:ss_calib_blocks}, but for the modified risk $\hat{R}_\text{PP}^{++}(q)$ from \eqref{eq:PP-risk_efficient}. In this case, the guarantee is only asymptotic, and thus we can estimate $\lambda^*$ with the same hold-out data used for computing the UCB. Figure~\ref{Fig:clt_comparison} compares the performance of RCPS with labeled data, the original CLT-based UCB with $\lambda=1$ (as in Figure~\ref{Fig:coverage_n_calib}), and the CLT-based UCB with the estimated $\lambda^*$. As can be seen, the latter indeed yields the least conservative prediction sets, as expected.

\begin{figure}[h]
    \centering
    \includegraphics[scale=0.32]{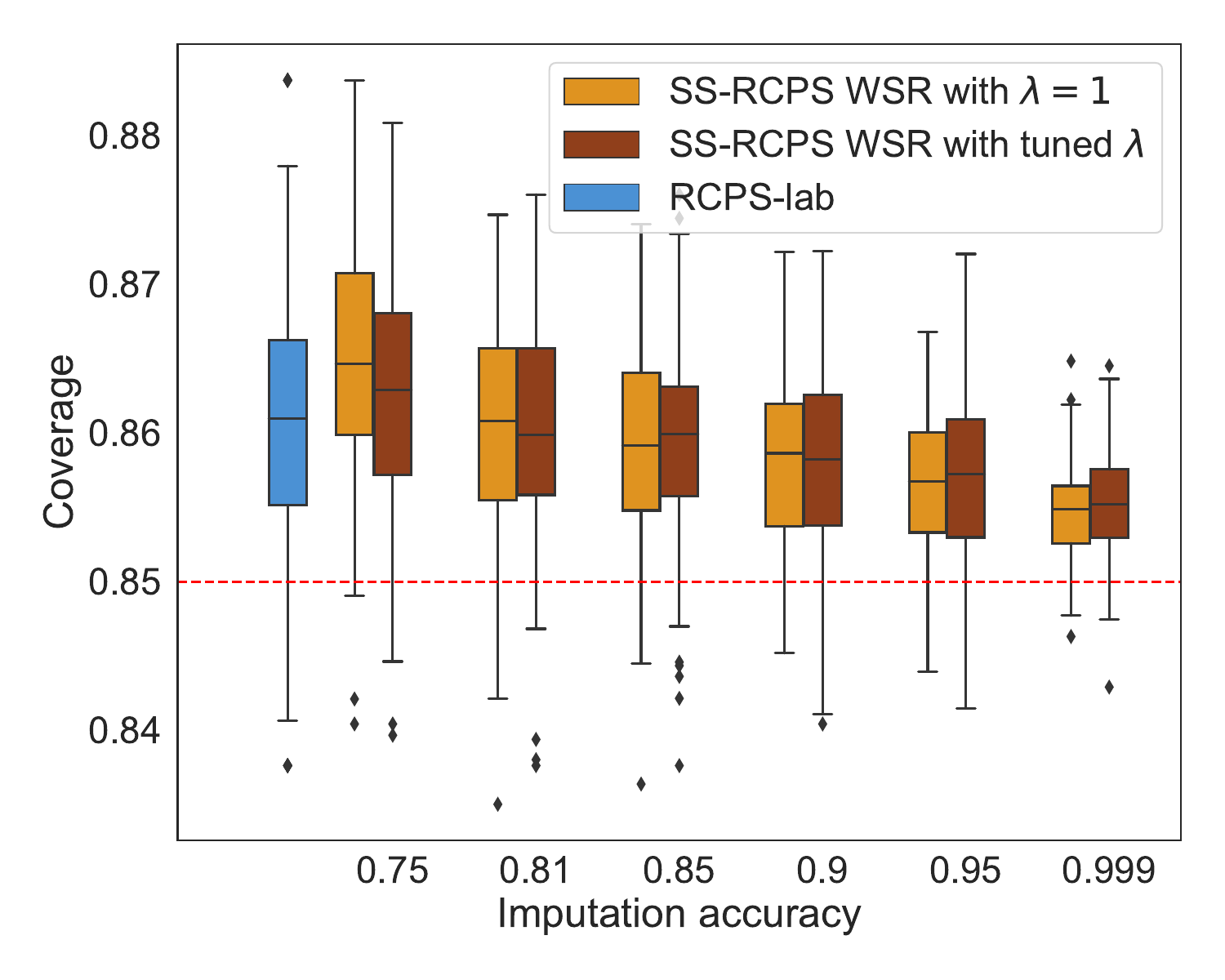}
    \caption{ImageNet classification experiment: the coverage rate of the constructed prediction sets as a function of the imputation accuracy, evaluated over 100 different random splits of the data. Comparison between RCPS using only labeled data, the original WSR UCB that uses $\lambda=1$, and the WSR UCB with the tuned $\lambda^*$. Other details are as in Figure~\ref{Fig:coverage_accuracy}.}
    \label{Fig:wsr_comparison_split}
\end{figure}

The second variant is the WSR-based UCB from Algorithm~\ref{alg:ss_calib_blocks}, which is supported by finite sample guarantees. Since the focus here is on non-asymptotic validity, estimating $\lambda^*$ requires additional data splitting. With limited labeled data, this is not desirable, as it makes the UCB more conservative due to the smaller sample size. However, with a relatively large amount of data, this method can be beneficial. 
Figure~\ref{Fig:wsr_comparison_split} compares RCPS using labeled data only, the WSR UCB with $\lambda=1$, and the WSR UCB with data splitting used to estimate $\lambda^*$. The setup uses 2000 labeled holdout points (200 for estimating $\lambda^*$ and the rest for UCB construction), and 23,000 unlabeled points (2300 for $\lambda^*$ estimation, and the remainder for UCB construction). Following that figure, we can see that for lower accuracies, the use of the estimated $\lambda^*$ is indeed beneficial: the estimated $\lambda^*$ effectively balances the reliance on unlabeled data. However, for higher accuracies, splitting the data slightly reduces statistical efficiency. This is because the optimal $\lambda^*$ is approximately $1$ as in the WSR UCB method from the main manuscript that uses a fixed $\lambda=1$, but the UCB of the tuned method is now constructed with fewer samples.

\subsection{ETSC Experiment: Additional Algorithmic and Experimental Details}
\label{app:NLP}

Here, we will describe the process for tuning $\underline{\hat{q}}$, where additional details can be found in~\cite{ringel2024early}. The algorithm divides the labeled calibration data into two disjoint sets of approximately the same size and takes a two-stage approach. Stage 1 utilizes the labeled validation set, $\mathcal{D}_{L,1}^{\text{cal}} = \{ (X_i, Y_i) \}_{i=1}^{m}$, to heuristically find a data-adaptive candidate threshold vector $\underline{\hat{\eta}}$. We begin by initializing $\hat{\underline{\eta}}$ with $\{\infty, \ldots, \infty\}$ which means there is no early stopping and the risk is controlled by definition. This vector is then updated in order, from the first timestep $t_1$ to the last $t_{\text{max}}$. For each $t$, the heuristic threshold $\underline{\hat{\eta}}_t$ is chosen such that the conditional empirical risk is approximately equal to $\alpha$. The detailed screening procedure of stage 1 is described in Algorithm~\ref{alg:candidate_screening}. 

In stage 2 the candidate threshold vector $\underline{\hat{\eta}}$ is used to form a stopping rule that satisfies~\eqref{eq:conditional_guarantee}. The idea is to use the labeled calibration set, $\mathcal{D}_{L,2}^{\text{cal}} = \{ (X_i, Y_i) \}_{i=1}^{n}$, to calibrate $\underline{\hat{q}}$ via RCPS with $\mathcal{Q}_{\text{grid}}$ that is determined by $\underline{\hat{\eta}}$.
Formally, we begin by defining:
\begin{align}
    &\hat{\underline{q}}_{t_{\text{max}}}=(\infty,\infty,\ldots,\hat{\eta}_{t_{\text{max}}}), \\
    &\hat{\underline{q}}_{t_{\text{max}}-1}=(\infty,\infty,\ldots,\hat{\eta}_{t_{\text{max}}-1},\hat{\eta}_{t_{\text{max}}}),\\
    \vdots\\
    &\hat{\underline{q}}_{1}=(\hat{\eta}_{1},\hat{\eta}_{2},\ldots,\hat{\eta}_{t_{\text{max}}-1},\hat{\eta}_{t_{\text{max}}}),
\end{align}
and $\mathcal{Q}_{\text{grid}}$ is constructed as follows:
$$\mathcal{Q}_{\text{grid}}=(\hat{\underline{q}}_{t_{\text{max}}},\hat{\underline{q}}_{t_{\text{max}}-1},\ldots,\hat{\underline{q}}_{1}).$$
The detailed procedure of stage 2 is provided in Algorithms~\ref{alg:valid_stage} and~\ref{alg:valid_stage_WSR}, which correspond to semi-supervised calibration for a binary loss and a general loss, respectively.

As a concluding remark, we provide some supplementary details regarding the ETSC experiment. We split the labeled calibration data in half for the first and second stages (150/2000 samples for each stage). The unlabeled calibration data includes 50,000 fixed samples for each experiment. 
We created timesteps by dividing each review into groups, ensuring each group contains approximately the same number of tokens.
\onecolumn

\begin{algorithm}[t]
\caption{Stage 1: Candidate Screening, adopted from~\cite{ringel2024early}}
\label{alg:candidate_screening}
\textbf{Input}: Calibration set $\mathcal{D}_{L,1}^{\text{cal}} = \{ (X_i, Y_i) \}_{i=1}^{m}$;
tolerable accuracy gap $\alpha$; grid resolution $\Delta$. \\
\textbf{Process:}\\
  \( \hat{\underline{\eta}} \leftarrow \{\infty, \ldots, \infty\} \) \\
  \For{\( t = 1, \ldots, t_{\textup{max}} \)} 
    {\( \underline{\eta} \leftarrow \underline{\hat{\eta}} \) \\
    \For{\( \xi = 0, \Delta, 2\Delta, \ldots, 1 \)}
      {\( \underline{\eta}_t \leftarrow \xi \) \\
      \( I \leftarrow \{i : \tau_{\underline{\eta}}(X_i) \leq t \}, \hspace{2mm} i\in\{1,\ldots,m\} \) {\color{gray}\Comment{Find samples with a halt time $\leq t$.}}\\
      \If{\( I = \emptyset \)}
        {Break inner loop and set $\hat{\underline{\eta}}_t \leftarrow \infty$ {\color{gray}\Comment{Cannot calculate the empirical risk}}
      }
      Calculate the empirical risk:\\
      \( \hat{R}_{\text{gap}}^{\leq t} \leftarrow \frac{1}{|I|}\sum_{i \in I} L_{\text{gap}}(Y_i,\hat{Y_i}^{\text{full}},\hat{Y_i}^{\text{early}}(\tau_{\underline{\eta}})) \) \\
      \If{\( \hat{R}_{\textup{gap}}^{\leq t} \leq \alpha \)}
        {
        Break inner loop and set  $\hat{\underline{\eta}}_t \leftarrow \xi$ {\color{gray}\Comment{Found the lowest $\underline{\eta}_t$ s.t. $\hat{R}_{\text{gap}}^{\leq t} \leq \alpha$.}} 
      }
    }
  }
\textbf{Output}: A heuristic candidate vector $\hat{\underline{\eta}}$
\end{algorithm}

\begin{algorithm}[t]
\caption{Stage 2: Calibration via Semi-Supervised RCPS for a Binary Loss}
\label{alg:valid_stage}
\textbf{Input}: Calibration set $\mathcal{D}_{L,2}^{\text{cal}} = \{ (X_i, Y_i) \}_{i=1}^{n}$;\\ 
unlabeled calibration data $\mathcal{D}_{U}^{\text{cal}}=\{X_j\}_{j=n+1}^{n+N}$; imputed labels $\{\tilde{Y}_j\}_{j=1}^{n+N}$; candidate thresholds $\hat{\underline{\eta}}$;\\
error levels $(\alpha,\delta_1,\delta_2)$; grid resolution $\Delta$. \\
\textbf{Process:}\\
\( \hat{\underline{q}} \leftarrow \{\infty, \ldots, \infty\} \) {\color{gray}\Comment Start with the most conservative stopping rule.}\\

\For{\( t = t_{\textup{max}}, t_{\textup{max}}-\Delta, \ldots, 1 \)} 
  {\( \underline{q} \leftarrow \hat{\underline{q}} \) {\color{gray}\Comment Gradually reveal another $\hat{\underline{\eta}}_t$ from the end and test it.} \\
  \( \underline{q}_t \leftarrow \hat{\underline{\eta}}_t \) \ \\
  \For{\( t' = t, \ldots, t_{\textup{max}} \)}
    {
    $I \leftarrow \{i : \tau_{\underline{q}}(X_i) \leq t' \}, \hspace{2mm} i\in \{1,\ldots,n\}$ {\color{gray}\Comment Find labeled samples with a halt time $\leq t'$.}\\
    \( J \leftarrow \{j : \tau_{\underline{q}}(X_j) \leq t' \}, \hspace{2mm} j\in \{n+1,\ldots,n+N\} \) {\color{gray}\Comment Find unlabeled samples with a halt time $\leq t'$.}\\
    \If{\( I = \emptyset \)}
        {
        Break both loops
    }
    Compute the unlabeled empirical risk: \( \hat{R}_{\text{U,gap}}^{\leq t'}(\underline{q}) = \frac{1}{|J|}\sum_{j \in J} L_{\text{gap}}(\tilde{Y}_j,\hat{Y_j}^{\text{full}},\hat{Y_j}^{\text{early}}(\tau_{\underline{q}})) \) \\
     Derive the exact UCB $\hat{R}^+_U(\underline{q})$ at level $\delta_1$ \\
    
    Compute the empirical clipped rectifying risk: \\
    \(\hat{R}_{\text{\text{rect},gap}}^{\leq t',\text{clip}}(\underline{q}) = \frac{1}{|I|}\sum_{j \in I} \left(L_{\text{gap}}(Y_i,\hat{Y_i}^{\text{full}},\hat{Y_j}^{\text{early}}(\tau_{\underline{q}}))- L_{\text{gap}}(\tilde{Y}_i,\hat{Y_i}^{\text{full}},\hat{Y_j}^{\text{early}}(\tau_{\underline{q}}))\right)_+\) \\
    Derive the exact UCB, $\hat{R}^{\text{clip}+}_{\text{rect}}(\underline{q})$ at level $\delta_2$ \\
    Derive $\hat{R}^+(\underline{q})=\hat{R}^+_U(\underline{q})+\hat{R}^{\text{clip}+}_{\text{rect}}(\underline{q})$ \\
    \eIf{$\hat{R}^+(\underline{q}) < \alpha$}
      {$\hat{\underline{q}} \leftarrow \underline{q}$
       }
       {Break both loops}
  }

}
\textbf{Output}: The calibrated parameter vector $ \hat{\underline{q}}$
\end{algorithm}

\begin{algorithm}[t]
\caption{Stage 2: Calibration via Semi-Supervised RCPS for a General Loss}
\label{alg:valid_stage_WSR}
\textbf{Input}: Calibration set $\mathcal{D}_{L,2}^{\text{cal}} = \{ (X_i, Y_i) \}_{i=1}^{n}$;\\ 
unlabeled calibration data $\mathcal{D}_{U}^{\text{cal}}=\{X_j\}_{j=n+1}^{n+N}$; imputed labels $\{\tilde{Y}_j\}_{j=1}^{n+N}$; candidate thresholds $\hat{\underline{\eta}}$;\\
error levels $(\alpha,\delta)$; grid resolution $\Delta$. \\

\algnewcommand{\LeftComment}[1]{{\color{gray}\(\triangleright\) #1}}

\textbf{Process:}\\
\( \hat{\underline{q}} \leftarrow \{\infty, \ldots, \infty\} \) {\color{gray}\Comment Start with the most conservative stopping rule.}\\

\For{\( t = t_{\textup{max}}, t_{\textup{max}}-\Delta, \ldots, 1 \)} 
  {\( \underline{q} \leftarrow \hat{\underline{q}} \) {\color{gray}\Comment Gradually reveal another $\hat{\underline{\eta}}_t$ from the end and test it.} \\
  \( \underline{q}_t \leftarrow \hat{\underline{\eta}}_t \) \ \\
  \For{\( t' = t, \ldots, t_{\text{max}} \)}
    {
    $I \leftarrow \{i : \tau_{\underline{q}}(X_i) \leq t' \}, \hspace{2mm} i\in \{1,\ldots,n\}$ {\color{gray}\Comment Find labeled samples with a halt time $\leq t'$.}\\
    \( J \leftarrow \{j : \tau_{\underline{q}}(X_j) \leq t' \}, \hspace{2mm} j\in \{n+1,\ldots,n+N\} \) {\color{gray}\Comment Find unlabeled samples with a halt time $\leq t'$.}\\
    \If{\( I = \emptyset \)}
        {
        Break both loops
    }

    For all $1\leq i\leq n$, compute\\
    $W_i=\frac{|I|}{|J|}\sum_{j=(i-1)\frac{|J|}{|I|}+n+1}^{i\frac{|J|}{|I|}+n}L_{\text{gap}}(\tilde{Y}_j,\hat{Y_j}^{\text{full}},\hat{Y_j}^{\text{early}}(\tau_{\underline{q}}))+L_{\text{gap}}(Y_i,\hat{Y_i}^{\text{full}},\hat{Y_i}^{\text{early}}(\tau_{\underline{q}}))-L_{\text{gap}}(\tilde{Y}_i,\hat{Y_i}^{\text{full}},\hat{Y_i}^{\text{early}}(\tau_{\underline{q}}))$ \\
        Derive a UCB $\hat{R}_{\text{PP}}^+(q)$ given $\hat{R}_\text{PP}(q)$ in \eqref{eq:ppi_decom} \\ \LeftComment{E.g., using WSR in Algorithm~\ref{alg:wsr} of Appendix~\ref{app:wsr}}\\
        
    
    \eIf{$\hat{R}_{\textup{PP}}^+(\underline{q}) < \alpha$}
      {$\hat{\underline{q}} \leftarrow \underline{q}$
       }
       {Break both loops}
  }

}
\textbf{Output}: The calibrated parameter vector $ \hat{\underline{q}}$
\end{algorithm}

\end{appendices}


\end{appendices}

\end{document}